\documentclass{article} 

\usepackage[table, dvipsnames, x11names]{xcolor}

\usepackage[accepted]{icml2023}

\usepackage{amsmath,amssymb}

\usepackage{booktabs}
\usepackage{siunitx}
\usepackage{makecell}
\usepackage{booktabs}
\usepackage{multirow}

\usepackage{graphicx}
\usepackage{tabularx}
\usepackage{wrapfig}
\usepackage{pgfplots}
\usepackage{tikz}
\usepackage{subcaption}
\usepackage[font={footnotesize}]{caption} %
\usepackage{tabularx}
\usepackage{textcomp}

\usepackage[utf8]{inputenc}
\usepackage{mathtools, nccmath}

\usepackage{enumitem} %

\usepackage[dvipsnames]{xcolor}
\usepackage{soul}

\usepackage[bookmarks=true, colorlinks=true,linkcolor=magenta,citecolor=RoyalBlue]{hyperref} 
\usepackage[nameinlink,capitalise]{cleveref}%
\usepackage{url}
\crefname{section}{§}{§§}
\Crefname{section}{§}{§§}

\usepackage[bb=boondox,bbscaled=.95]{mathalfa}
\usepackage{amsmath, amssymb, amsfonts, amsthm, bm}
\usepackage{physics}
\usepackage{cancel}
\usepackage{thmtools, thm-restate}

\usepackage{enumitem} 
\declaretheorem{theorem}

\newtheorem{definition}{Definition}

\definecolor{bg}{gray}{0.95}
\usepackage{xcolor}  
\usepackage[dvipsnames]{xcolor} 

\usepackage{amsmath,amsfonts,bm}

\def\eqref#1{equation~\ref{#1}}

\def\1{\bm{1}}

\DeclareMathAlphabet{\mathsfit}{\encodingdefault}{\sfdefault}{m}{sl}
\SetMathAlphabet{\mathsfit}{bold}{\encodingdefault}{\sfdefault}{bx}{n}

\DeclareMathOperator*{\argmax}{arg\,max}

\def\xunderbrace#1_#2{{\underbrace{#1}_{#2}}}
\def\xoverbrace#1^#2{{\overbrace{#1}^{#2}}}

\definecolor{olive}{rgb}{0.6, 0.6, 0.2}
\definecolor{sand}{rgb}{0.8666666666666667, 0.8, 0.4666666666666667}
\definecolor{wine}{rgb}{0.5333333333333333, 0.13333333333333333, 0.3333333333333333}
\definecolor{deblue}{RGB}{11,132,147}
\definecolor{ocra}{RGB}{204, 119, 34}

\newcommand{\fcircle}[2][red,fill=red]{\tikz[baseline=-0.5ex]\draw[#1,radius=#2] (0,0.03) circle ;} 

\newcommand{\ourmethod}{\textsc{DevFormer}} 

\def \papertitle{DevFormer: A Symmetric Transformer for Context-Aware Device Placement}

\icmltitlerunning{\papertitle}

\icmlsetsymbol{equal}{*}

\begin{document}
 

\twocolumn[
\begin{icmlauthorlist}

\icmltitle{\papertitle}

\icmlauthor{Haeyeon Kim\textsuperscript{*}}{kaist_ee}  
\icmlauthor{Minsu Kim\textsuperscript{*}}{kaist} 
\icmlauthor{Federico Berto}{kaist} 
\icmlauthor{Joungho Kim}{kaist_ee}
\icmlauthor{Jinkyoo Park}{kaist}

\end{icmlauthorlist}

\icmlaffiliation{kaist_ee}{Department of Electrical Engineering, Korea Advanced Institute of Science and Technology (KAIST)}
\icmlaffiliation{kaist}{Department of Industrial and Systems Engineering, Korea Advanced Institute of Science and Technology (KAIST)}

\icmlcorrespondingauthor{Jinkyoo Park}{jinkyoo.park@kaist.ac.kr}

\icmlkeywords{Machine Learning, ICML}

\vskip 0.3in
]

\printAffiliationsAndNotice{\icmlEqualContribution} 
\sethlcolor {Aquamarine}

\begin{abstract}

In this paper, we present \ourmethod{}, a novel transformer-based architecture for addressing the complex and computationally demanding problem of hardware design optimization. Despite the demonstrated efficacy of transformers in domains including natural language processing and computer vision, their use in hardware design has been limited by the scarcity of offline data. Our approach addresses this limitation by introducing strong inductive biases such as relative positional embeddings and action-permutation symmetricity that effectively capture the hardware context and enable efficient design optimization with limited offline data.  We apply \ourmethod{} to the problem of decoupling capacitor placement and show that it outperforms state-of-the-art methods in both simulated and real hardware, leading to improved performances while reducing the number of components by more than $30\%$. Finally, we show that our approach achieves promising results in other offline contextual learning-based combinatorial optimization tasks.
\end{abstract}

\section{Introduction}

The development of artificial intelligence (AI) has been greatly facilitated by advancements in high-performance computing systems. However, as the need for faster data processing grows with recent advances in AI architecture scaling, the complexity of hardware design is increasing. As a result, human experts are no longer able to design hardware without the aid of electrical design automation (EDA) tools; despite their utility, the use of these tools is often hindered by long simulation times and insufficient computing resources, making the application of machine learning (ML) techniques in hardware design more and more crucial.

Recent studies have shown the potential of deep reinforcement learning (DRL) for 
 sequential decision-making in various tasks in chip design, including chip placement \citep{nature, vlsi_placement}, routing \citep{Liao2019ADR,liao2020track}, circuit design \citep{sizing}, logic synthesis \citep{logic1, logic2} and bi-level hardware optimization \citep{nips_joint}. However, these methods have limitations. Firstly, they rely on online simulators, which are both slow and inaccurate. As a result, learning with offline expert data is more reliable, but such data is limited. Secondly, hardware design involves multi-level problems that change depending on the context, making it necessary to have a policy that can generalize to new problem contexts.
 
The Transformer has been recognized as a promising architecture for contextual models, owing to its capability to process sequential data in parallel, handle long-term dependencies and exhibit high expressivity \citep{transformer}. This has led to their widespread adoption in a variety of domains, including natural language processing (NLP) \citep{gpt3}, computer vision \citep{ViT}, graph representation learning \citep{graphormer}, combinatorial optimization \citep{kool_attention} and reinforcement learning \citep{chen2021decision}. However, deriving an effective offline contextual policy with the Transformer requires a huge amount of offline data covering the broad context and input data space, which is often infeasible in the semiconductor industry.

\begin{figure}[t]
    \centering \includegraphics[width=.9\linewidth]{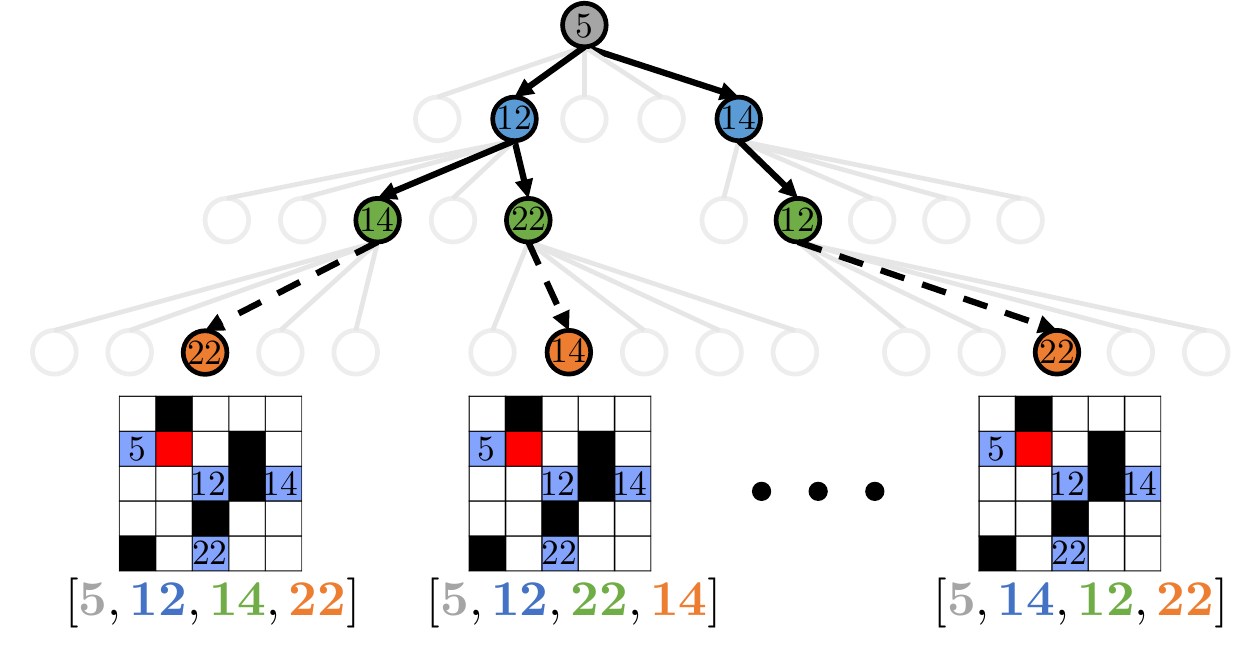}
    \caption{Example of action-permutation (AP) symmetric solutions: different action trajectories lead to the same solution group.}
    \label{fig:symmetricity}
    \vspace{-2mm}
\end{figure}

Training a flexible contextual design policy based on Transformer with a limited volume of offline data is a challenging task. The current study aims to overcome this issue by utilizing inductive biases for the target hardware design problem. For example, the device placement problem has an input order invariance property in that input permutations do not affect the output, while Transformers have a strong order bias. \cref{fig:symmetricity} illustrates this property: for example, $\{1,2,3,4\}$ and $\{2,3,1,4\}$ are considered action permutation (AP) symmetric solutions. In addition, the relative locations of hardware devices matter more than their absolute positions, while Transformer heavily uses an absolute positional encoding. Thus, we can exploit such properties of the target problem to modify Transformer architecture for deriving an effective contextual policy with a limited offline dataset.

In this paper, we present \ourmethod{}, a novel transformer model for solving contextual offline hardware design problems. By incorporating strong domain-specific inductive biases, our model learns more efficiently and effectively, overcoming the limitations of traditional transformer architectures. We demonstrate the proposed approach on a novel hardware benchmark and validate its applicability on a real-world high bandwidth memory.

\vspace{3mm}
We summarize our contributions as follows:
\smallskip
\vspace{-4mm}
\begin{itemize}
    \item We propose a new positional embedding technique for the encoder of the transformer, utilizing relative chip location, and a probing-port context network (PCN) for the decoder. To address the issue of order bias in traditional positional embedding, we propose a recurrent context network (RCN) that only references the previously selected node. The PCN, RCN, and encoded node embeddings are used in an attentive manner to generate actions during the decoding process.
    \smallskip
    \vspace{-3mm}
    \item To address order biases and impose AP symmetricity, we introduce a novel regularization loss term for the placement problem symmetricity. This approach can also be applied to other sequential design methods, such as DRL, with similar order bias issues.
    \smallskip
    \item We demonstrate our approach to the decap placement problem (DPP) and release the novel DPP benchmark for researchers to evaluate and improve ML methodologies for hardware design challenges. As a means of promoting transparency and reproducibility, we make the source codes of our method and the baselines discussed in this paper publicly available online\footnote{\href{https://github.com/kaist-silab/devformer}{https://github.com/kaist-silab/devformer}} as well as an accompanying interactive demonstration program\footnote{\href{https://dppbench.streamlit.app}{https://dppbench.streamlit.app}} to facilitate engagement and experimentation.
    
    \vspace{30mm}
\end{itemize}

\begin{figure}[t]
    \centering
    \includegraphics[width=0.48\textwidth]{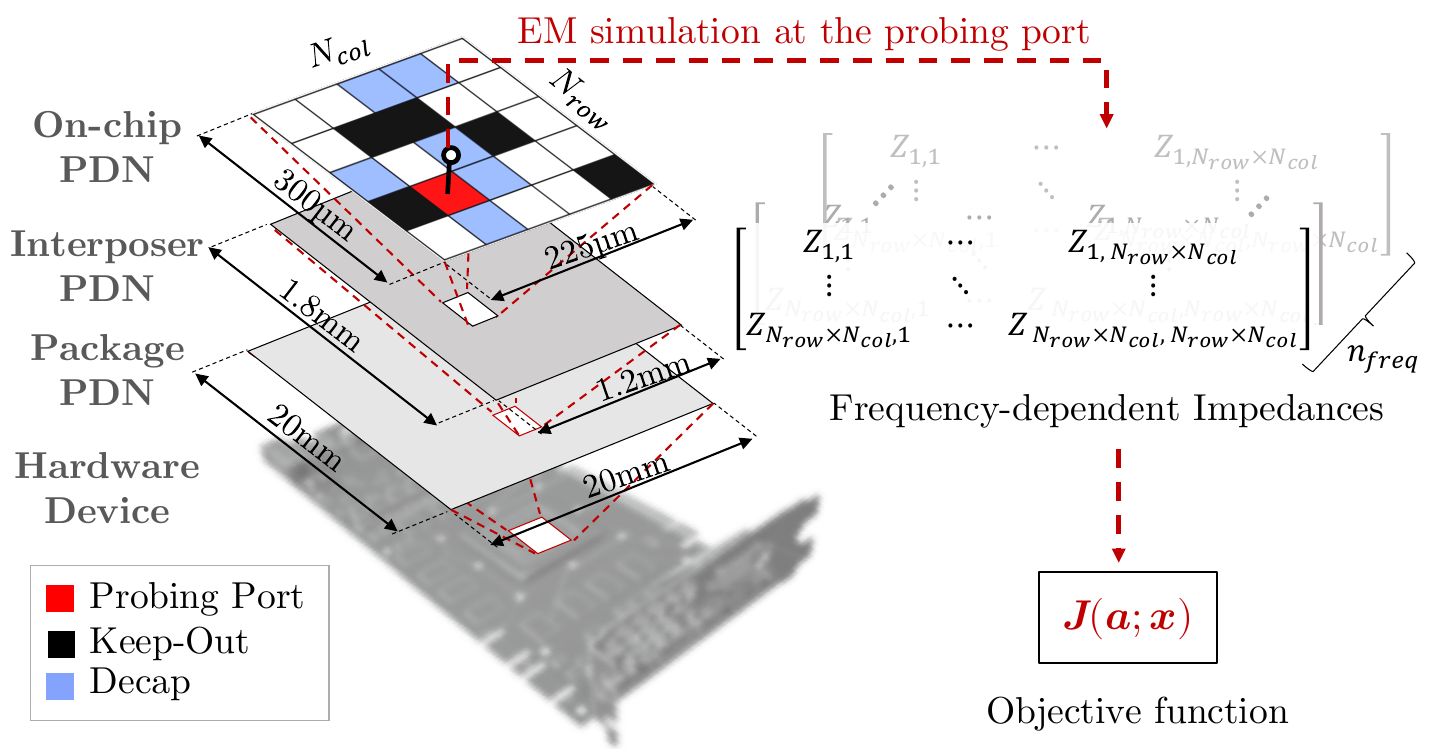}
    \vspace{-3mm}
    \caption{Grid representation of the target on-chip PDN of a hardware device and the EM simulation at the probing port to evaluate the objective function. }
    \label{figure:DPP(a)}
    \vspace{-3mm}
    \label{fig:DPP}\
\end{figure} 
\section{Preliminaries and Background}

\subsection{Decap Placement Problem (DPP)}
This paper seeks to solve the decap placement problem (DPP), one of the essential hardware design problems. A decoupling capacitor (decap) is a hardware component that reduces power noise along the power distribution network (PDN) of hardware devices such as CPUs, GPUs, and AI accelerators, and improves the power integrity (PI). 
 
Formally, DPP is a black-box contextual optimization problem to find the optimal placement of decap $\bm{a}=\{\bm{a}_1,...,\bm{a}_K\}$ that maximizes the PI objective $\mathcal{J}(;\bm{x})$. Note the objective is contextualized by the target hardware feature vectors $\bm{x}$ with the constraint of a limited number of decap $K$. Our research aims to solve:
\begin{align*}
    &\bm{a^*} = \arg\max_{a} \mathcal{J}(\bm{a};\bm{x})
    \\
    &\text{s.t} \quad K \leq K^{*}, x \in \mathcal{X}
\end{align*}
$\mathcal{X}$ refers to the context space of the target hardware. Note that the number of decap $K$ is a crucial budget for DPP optimization as placing a decap is costly in semiconductor industries \cite{decap_cost}.

\subsection{Contextual Markov Decision Processes (cMDP)}
\vspace{-2mm}
\label{cMDP}

We formulate DPP as a contextual Markov decision process \citep[cMDP]{hallak2015contextual} to decompose the joint action policy into a sequential component policy to overcome the high dimensionality issue.
Specifically, our objective function $\mathcal{J}(;\boldsymbol{x})$ is determined by the PDN, which is contextualized by $\boldsymbol{x}$. The contextualized PDN is represented as a set of three-dimensional feature vectors $\boldsymbol{x} = {\{\boldsymbol{x}_i\}}_{i=1}^{N_{\text{row}} \times N_{\text{col}}}$, where each grid (i.e., port) on the PDN is represented as $\boldsymbol{x}_i={(x_i, y_i, c_i)}$, in which $x_i,y_i$ indicate the 2D coordinates of the location; $c_i \in \{0,1,2\}$ indicates the condition of the port, whether it belongs to a probing port, keep-out regions, or decap allowed ports, respectively. See \cref{append:data-structure} for further details.


\begin{figure*}[t]
    \centering
    \begin{subfigure}[b]{.75\textwidth}
    \includegraphics[width=\textwidth]{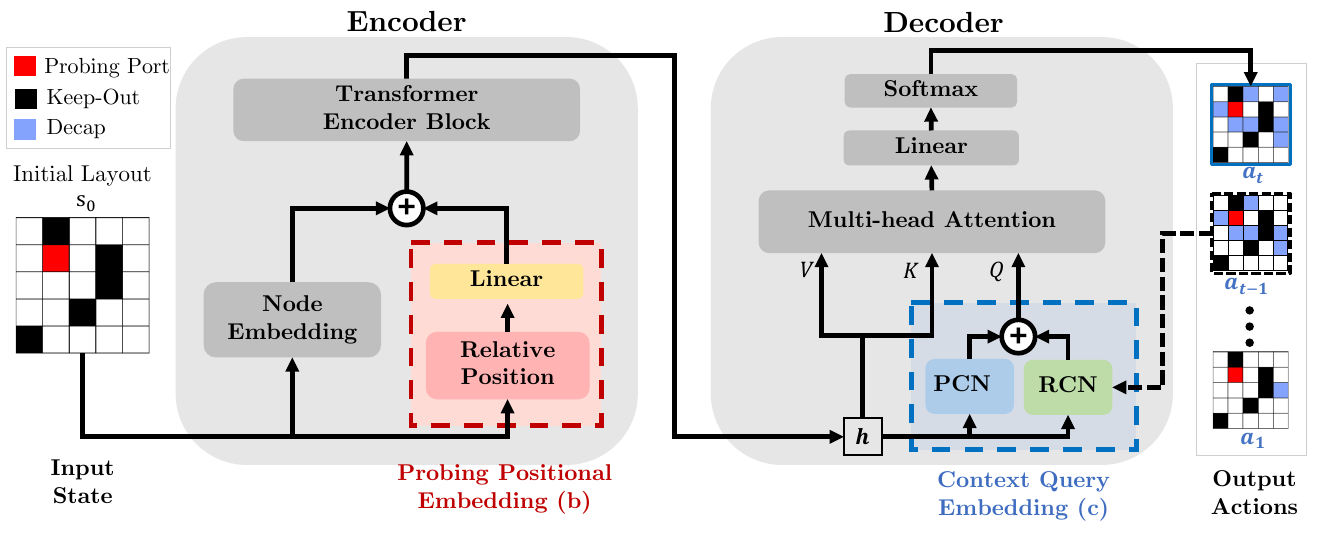}
    \vspace{-5mm}
    \caption{Overall \ourmethod{} architecture}
    \label{fig:main_a}
    \end{subfigure}\qquad
\begin{subfigure}[b]{.2\textwidth}
    \centering
    \includegraphics[width=.9\textwidth]{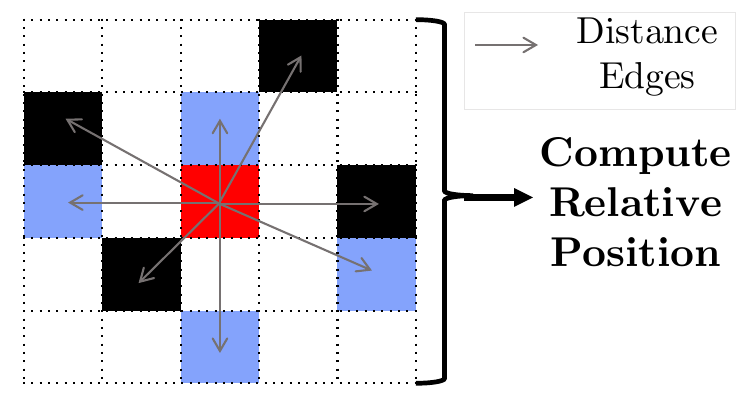}
    \vspace{-1mm}
    \caption{Probing Positional Embedding}    
    \label{fig:main_b}
    \vspace{0mm}
    \includegraphics[width=.9\textwidth]{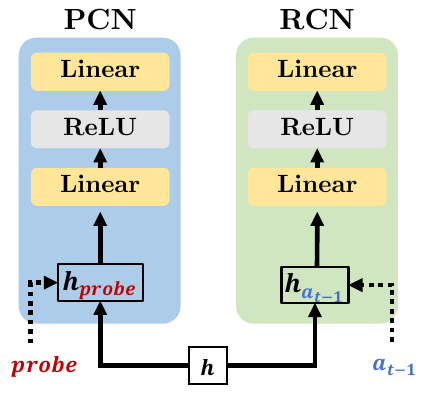}
    \vspace{-2mm}
    \caption{Context Query Embedding}
    \label{fig:main_c}
\end{subfigure}
\vspace{-2mm}
\caption{ (a) Given design constraints and previous actions, the \ourmethod{} model generates the optimal device placement action $a_t$. (b)  Probing Positional Embedding (\fcircle[fill=BrickRed!80!white]{3pt}) uses relative positions with respect to the probing port. (c)  PCN (probing port context network \fcircle[fill=RoyalBlue!80!white]{3pt}) and  RCN (recurrent context network \fcircle[fill=OliveGreen!80!white]{3pt}) use embedding $h$ to capture contextual information of initial design conditions and stages of the partial solution, respectively.}
\label{fig:main}
\vspace{-4mm}
\end{figure*}



The design process sequentially places a pre-defined number $K$ of decaps on the available PDN ports. We model this cMDP with state, action, and policy as follows.

\textbf{State} $\boldsymbol{s}_{t}$ contains the task-condition (i.e., problem context vectors) $\boldsymbol{x}$ and the previous selected actions: $\boldsymbol{s}_{t} = \{\boldsymbol{x},\boldsymbol{a}_{1:t-1}$\}.

\textbf{Action} $a_{t} \in \{1,...,N_{\text{row}} \times N_{\text{col}} \} \setminus \boldsymbol{s}_{t-1}$ is defined as an allocation of a decap to one of the available ports on PDN. The available ports are the ports on PDN, except for the probing port, keep-out, and previously selected ports. The concatenation of sequentially selected actions $\boldsymbol{a}=a_{1:K}$ indicates the final decap placement \textit{solution}.

\textbf{Policy} $\pi_{\theta}(\boldsymbol{a}|\boldsymbol{x})$ is the probability of producing a specific solution $\boldsymbol{a}=a_{1:K}$, given the problem context vectors $\boldsymbol{x}$, and is factorized as:
\vspace{-4mm}
\begin{equation}
    \pi_{\theta}(\boldsymbol{a}| \boldsymbol{x}) = \prod_{t=1}^{K} p_{\theta}(a_t | \boldsymbol{s_{t}}),
\label{policy}
\vspace{-2mm}
\end{equation}
where $p_{\theta}(a_t | \boldsymbol{s_{t}})$ is the segmented one-step action policy parameterized by the neural network.

The objective of DPP is to find the optimal parameter $\theta^{*}$ of the policy $\pi_{\theta}(\cdot|\boldsymbol{x})$ as:
\vspace{-1mm}
\begin{align}
    {\theta}^*= \argmax_\theta{\mathbb{E}}_{\boldsymbol{x} \sim p_{\mathcal{X}}(\boldsymbol{x})}{\mathbb{E}}_{\textcolor{black}{\boldsymbol{a} \sim \pi_\theta(\cdot|\boldsymbol{x})}}\big[\mathcal{J}(\boldsymbol{a};\boldsymbol{x} )\big], 
\vspace{-2mm}
\end{align}
where $p_{\mathcal{X}}(\boldsymbol{x})$ is the probability distribution for varying \textit{task-condition} $\boldsymbol{x}$ and $\mathcal{J}$ is objective function. Finding the optimal policy for various DPPs with changing conditions is a contextual learning problem, in which each DPP has a distinct context. Once the task $\boldsymbol{x}$ is sampled by $p_{\mathcal{X}}(\boldsymbol{x})$, the state-action space with complexity of ${N_{\text{row}}\times N_{\text{col}} \choose {K}}$ is determined. Then, an efficient policy $\pi_{\theta}(\boldsymbol{a}|\boldsymbol{x})$ should capture the contextual features among varying task conditions $\boldsymbol{x}$.

Note that DPP is an episodic task, where the reward is defined as the objective of the final state solution; the reward is the same as objective $\mathcal{J}$.

\subsection{Objective Function $\mathcal{J}$}

The objective function is a black-box simulator based on the electrical magnetic (EM) simulation or lab-level electrical measurement on fabricated products. To benchmark DPP as an ML task, we implemented the objective function with approximated modeling of electrical components as the fast Python simulator. As shown in \cref{fig:DPP}, calculating objective function requires frequency-dependent impedances calculated through EM simulation at the probing port, and such process is costly in terms of time and computation. See \cref{append:objective} for a detailed description of our modeling of the objective function.

\section{Methodology}

\begin{figure*}[h]
\centering
\includegraphics[width=0.95\textwidth]{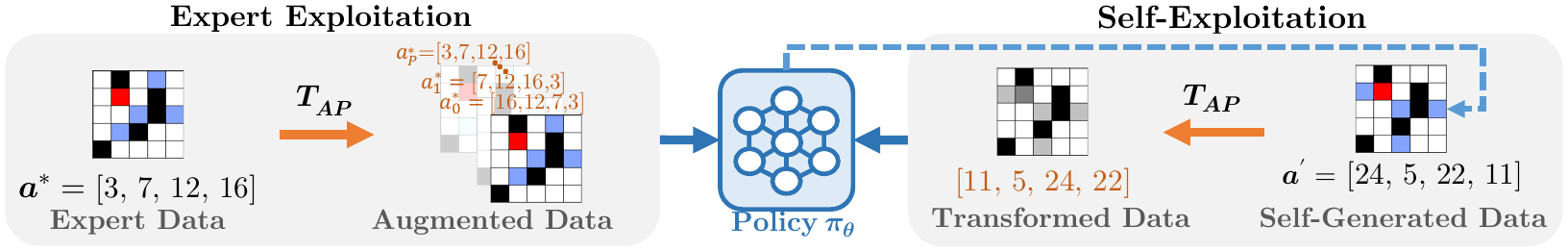}
\vspace{-1mm}
\caption{Collaborative symmetricity exploitation (CSE) framework with \ourmethod{} policy. Training is carried out via expert exploitation (EE), which employs offline high-quality data, while self-exploitation (SE) uses data generated by a copy of the policy itself. AP-symmetricity is imposed through the $T_{AP}$ transformation on both expert and self-generated data.}
\vspace{-2mm}
\label{fig:CSE}
\end{figure*}

\subsection{\ourmethod{} Architecture}

\ourmethod{}, illustrated in \cref{fig:main}, is a novel transformer-based architecture that incorporates hardware-aware prior components, such as the  probing port positional embedding (\fcircle[fill=BrickRed!80!white]{3pt}) for the encoder and the  probing port context network (\fcircle[fill=RoyalBlue!80!white]{3pt}) and recurrent context network (\fcircle[fill=OliveGreen!80!white]{3pt}) for the decoder. The encoder, $f_{\text{enc}}$, maps the task condition $\bm{x}$ to a high-dimensional embedding $\bm{h}$, through the function $f_{\text{enc}}(\bm{x})$. Similarly to the decoding scheme in pointer network (PointerNet) \citep{vinyals_pointer} and attention model (AM) \citep{kool_attention}, our decoder $p_{\text{dec}}(\bm{a}_t|\bm{h},\bm{a}_{t-1})$ samples the indices of the placements in an auto-regressive manner: $\bm{a}_{t} \sim p_{\text{dec}}(\bm{a}_t|\bm{h},\bm{a}_{t-1})$.

\paragraph{Encoder}Our encoder consists of $L$ layers of multi-head attention (MHA) and feedforward (FF), akin to the transformer network proposed in \citet{transformer} as illustrated in \cref{fig:main_a}. Before processing by MHA and FF, the task condition $\bm{x}$ is processed by two networks: node embedding and probing port positional embedding (PPE \fcircle[fill=BrickRed!80!white]{3pt}). The node embedding is the node-wise linear embedding of $\bm{x}$: $\phi_{node}(\bm{x})$. The PPE is the linear projection $\phi_{\text{PPE}}$ of relative positions with respect to the probing port node $\bm{x}_{\text{probe}}$: $\phi_{\text{PPE}}\left(\{||\bm{x}_i-\bm{x}_{\text{probe}}||\}_{i=1}^{N}\right)$ as shown in \cref{fig:main_b}.  
The sum of processed node embeddings and PPE is then fed into the L-layered MHA and FF to generate $\bm{h}$\footnote{Note that we indicate $\bm{h}^L$ as a high dimensional embedding $\bm{h}$ for a simple notation.}:
\begin{align*}
 &\bm{h}^0 \leftarrow \phi_{\text{node}}(\bm{x}) + \phi_{\text{PPE}}(\{||\bm{x}_i-\bm{x}_{\text{probe}}||\}_{i=1}^{N})\\
 &\bm{h}^{l+1} \leftarrow \text{BN}\left(\Phi_{\text{FF}}\left(\text{BN}\left(\Phi_{\text{MHA}}\left(\bm{h}^l\right)\right)\right)\right)  
\end{align*}
where $\text{BN}$ stands for batch normalization.

\paragraph{Decoder} The decoder $p_{\text{dec}}(\bm{a}_t|\bm{h},\bm{a}_{t-1})$ is composed of a single layer MHA to generate the action probability distribution with SoftMax. We improve generalization capabilities by capturing contextual information: we generate the query key $Q$ via a context query embedding divided into a probing port context network (PCN \fcircle[fill=RoyalBlue!80!white]{3pt}) and recurrent context network (RCN \fcircle[fill=OliveGreen!80!white]{3pt}), as shown in \cref{fig:main_c}. The PCN focuses on processing the embedding of probing port $\bm{h}_{\text{probe}}$ where RCN recurrently focuses on the embedding of previously selected action $\bm{h}_{\bm{a}_{t-1}}$: 
\begin{align*}
  &\phi_{\text{PCN}}(\bm{h}_{\text{probe}}) = \text{MLP}(\bm{h}_{\text{probe}})\\
  &\phi_{\text{RCN}}(\bm{h}_{\bm{a}_{t-1}}) = \text{MLP}(\bm{h}_{\bm{a}_{t-1}})\\
  Q = &\phi_{\text{Linear}}(\phi_{\text{PCN}}(\bm{h}_{\text{probe}}) + \phi_{\text{RCN}}(\bm{h}_{\bm{a}_{t-1}}))
\end{align*}
Then, the probability distribution is computed with MHA and linear projection $\phi_{\text{Linear}}$:
\begin{align*}
p_{\text{dec}}(\bm{a}_t|\bm{h},\bm{a}_{t-1}) = \text{SoftMax}\left(\phi_{\text{Linear}}\left(\left(\text{MHA}\left(Q,K,V\right)\right)\right)\right)
\end{align*}
where $K$ and $V$ are a linear projection of $\bm{h}$.

\subsection{Action-permutation Symmetricity and Order Bias}
The symmetricity found in placement problems is the action-permutation (AP)-symmetricity, i.e., placement order does not affect the design performance. Let us denote $t_i$ as a permutation of an action sequence $\{1,...,K\}$, where $K$ is the length of the action sequence. We then define the AP-transformation $T_{\text{AP}}=\{t_i\}_{i=1}^{K!}$ as a set of all possible permutations. The AP-symmetricity of DPP is the property that we want to impose on the design solver.
\vspace{2mm}

\begin{definition}[\textbf{AP-symmetricity}] For any $\boldsymbol{a} \in \mathcal{A}$, $\boldsymbol{x} \in \mathcal{X}$, $t \in T_{AP}$,  the following holds:
\vspace{-2mm}
\begin{itemize}
    \item A scalar-valued function $f:\mathcal{A} \times \mathcal{X}  \rightarrow \mathbb{R} $ is AP-symmetric if $f(\boldsymbol{a},\boldsymbol{x}) = f(t(\boldsymbol{a}),\boldsymbol{x})$. 
    \vspace{-1mm}
    \item A conditional probability distribution $\pi$ is AP-symmetric if $\pi(\boldsymbol{a}|\boldsymbol{x}) = \pi(t(\boldsymbol{a})|\boldsymbol{x})$.
\end{itemize}
\label{def:ap-sym}
\vspace{-2mm}
\end{definition}
$\mathcal{A}$ is the solution space and $\mathcal{X}$ is the task-condition space.

The objective function $\mathcal{J}:\mathcal{A} \times \mathcal{X} \rightarrow \mathbb{R}$ of DPP is an AP-symmetric function because $\boldsymbol{a}$ and $t(\boldsymbol{a})$ have identical placement design. AP-symmetricity can thus be induced to the policy $\pi$ (conditional probability) to reflect the AP-symmetricity of an objective function $\mathcal{J}$.
Moreover, we define an order bias metric, $b(\pi;\boldsymbol{p})$, to measure the AP-symmetricity.

\vspace{2mm}

\begin{definition}[\textbf{Order bias on distributions $\boldsymbol{p}$}] For a conditional probability $\pi(\boldsymbol{a}|\boldsymbol{x})$, where $\boldsymbol{x} \in \mathcal{X}$ and $\boldsymbol{a} \in \mathcal{A}$, the order bias $b(\pi;\boldsymbol{p})$, where $\boldsymbol{p}$ = $\{p_{\mathcal{X}}, p_{\mathcal{A}}, p_{T_{AP}}\}$, refers to: 
\vspace{-1mm}
\begin{equation*}
    b(\pi;\boldsymbol{p}) = \mathbb{E}_{p_{\mathcal{X}}(x)}\mathbb{E}_{p_{\mathcal{A}}(\boldsymbol{a})}\mathbb{E}_{p_{T_{\text{AP}}}(t)}[||\pi(\boldsymbol{a}|\boldsymbol{x})-\pi(t(\boldsymbol{a})|\boldsymbol{x})||_{1}]
\end{equation*}
\label{def:order-bias}
\vspace{-6mm}
\end{definition}
Intuitively, the order bias $b(\pi;\boldsymbol{p})$ is a general property of a sequential solution generation scheme. It measures how much the solver $\pi(\boldsymbol{a}|\boldsymbol{x})$ has different probabilities of generating AP-symmetric solutions. The order bias metric holds for the following theorem:

{\begin{theorem} A task-conditioned policy $\pi(\boldsymbol{a}|\boldsymbol{x})$ is AP-symmetric if and only if its order bias is zero ($b(\pi;\boldsymbol{p})=0$) while the distributions are non-zero, $p_{\mathcal{X}}(\boldsymbol{x})>0, p_{\mathcal{A}}(\boldsymbol{a})>0, p_{T_{AP}}(t)>0$, for any $x \in \mathcal{X}$, $\boldsymbol{a} \in \mathcal{A}$ and $t \in T_{AP}$.
\label{thm:order-bias}
\end{theorem}
We report the detailed proof of \cref{thm:order-bias} in \cref{append:proof}.

\subsection{Collaborative Symmetricity Exploitation (CSE)}

To induce the AP-symmetricity to the trained \ourmethod{}, and thus to improve its generalization capability and data efficiency in training with low-data regimes, we design a novel training framework called \textit{collaborative symmetricity exploitation} (CSE). \cref{fig:CSE} shows the overall scheme: CSE exploits symmetricity in both expert and self-generated data.

\paragraph{Expert Exploitation} The major role of expert exploitation is to train high-quality symmetric contextualized policies for various task conditions \textbf{$x$} by leveraging offline expert data \textbf{$a^{*}$} from the offline expert dataset $D_{\text{exp}} = \{(x^{(i)},a^{(i)*})\}_{i=1}^{N}$ with $T_{\text{AP}}$. $T_{\text{AP}}$ transforms each offline expert data \textbf{$a^{*}$} for $P$ times to \textit{augment} the offline expert dataset to reflect the AP-symmetric nature of the placement task. Specifically, we randomly choose $\{t_1,...,t_P\} \subset T_{\text{AP}}$ to generate $D_{\text{aug}} = \{\big(x^{(i)},a^{(i)*}\big),\big(x^{(i)},t_{1}(a^{(i)*})\big),...,\big(x^{(i)},t_{P}(a^{(i)*})\big)\}_{i=1}^{N}$. Then, $\mathcal{L}_{\text{Expert}}$ is expressed as a \textit{teacher-forcing} imitation learning scheme with the augmented expert dataset $D_{aug}$. 
\begin{equation}
\vspace{-1mm}
\mathcal{L}_{\text{Expert}} = -\mathbb{E}_{\boldsymbol{a}^{*}, \boldsymbol{x} \sim D_{\text{aug}}}[log \pi_{\theta}(\boldsymbol{a}^{*}|\boldsymbol{x})] \\
\end{equation}
Note that expert exploitation is expected to reduce order bias defined with the three uniform distributions; $\boldsymbol{x}$ of $\mathcal{U}_{D_{\text{exp}}}(\boldsymbol{x})$, $\boldsymbol{a}$ of $\mathcal{U}_{D_{\text{exp}}}(\boldsymbol{a})$, and $t$ of $\mathcal{U}_{T_{\text{AP}}(t)}$.

\paragraph{Self-Exploitation} While $D_{\text{aug}}$ only contains expert quality data, self-exploitation uses self-generated data, whose quality is poor initially but improves over the training phase. Thus, the self-exploitation scheme is designed to induce the AP-symmetricity in a wider action space to achieve greater generalization capability. 
\begin{equation}
\vspace{-2mm}
\mathcal{L}_{\text{Self}} = \mathbb{E}_{\mathcal{U}_{\mathcal{X}}(\bm{x})}\mathbb{E}_{\pi_{\tilde{\theta}}(\cdot|\boldsymbol{x})}\mathbb{E}_{\mathcal{U}_{T_{\text{AP}}}(\bm{t})}[||\pi_{\tilde{\theta}}(\boldsymbol{a'}|\boldsymbol{x})-\pi_{\theta}(t(\boldsymbol{a'})|\boldsymbol{x})||_{1}]
\end{equation}

Formally, the self-exploitation loss is a special form of order bias defined based on the distributions, $\boldsymbol{x} \sim \mathcal{U}_{\mathcal{X}}$, $\boldsymbol{a} \sim \pi_{\tilde{\theta}}$ (current policy) and $t \sim \mathcal{U}_{T_{\text{AP}}}$, where $\mathcal{U}$ is a uniform distribution; $\mathcal{L}_{\text{Self}} = b(\pi_{\theta},\boldsymbol{p}=\{\mathcal{U}_{\mathcal{X}},\pi_{\tilde{\theta}},\mathcal{U}_{T_{\text{AP}}}\})$.

\paragraph{Loss function}

We design the \ourmethod{} loss term $\mathcal{L}$ consisting of expert exploitation loss $\mathcal{L}_{\text{Expert}}$ and self-exploitation loss $\mathcal{L}_{\text{Self}}$ to reduce order bias (\cref{def:order-bias}): 
\begin{align}
\vspace{-3mm}
&\mathcal{L} := \mathcal{L}_{\text{Expert}} + \lambda \mathcal{L}_{\text{Self}}
\label{L}
\vspace{-4mm}
\end{align}
where $\lambda$ is a hyperparameter that adjusts the weighted ratio between $\mathcal{L}_{\text{Expert}}$ and $\mathcal{L}_{\text{Self}}$.

\section{Experimental Results}

\subsection{Dataset and Benchmark}
\label{4.1}
\paragraph{Benchmark Description}

The DPP is an important design optimization task to improve the power integrity performance of hardware to find optimal design $\bm{a}$ of context $\bm{x}$ that maximizes objective value $\mathcal{J}(\bm{a};\bm{x})$ with a limited number of decaps $K$. In this benchmark (1) we measure objective score with $K=20$ and (2) we measure minimum $K$ for satisfying target objective $J=J^{*}$. Also, we evaluate the sample efficiency both for training time and test time. For the training time efficiency, we measure the number of offline datasets $N$. For the test time efficiency, the number of simulation shots $M$.   

\paragraph{Baselines} We report various existing baselines for DPP devised by hardware domain researchers. We categorize them into four categories:
\begin{itemize}
\setlength\itemsep{1em}
    \item \textbf{Test Time Search.} A traditional search method of random search and genetic algorithm (GA) on the test time are benchmarked.
    \item \textbf{Online Test Time Adaptation.} Online DRL methods for DPP that directly solve test problems with the learning procedure are reported: CNN deep Q learning \citep[CNN-DQN]{deepQ_hyunwook}, CNN double DQN \citep[CNN-DDQN]{doubleQ_lin}, pointer network policy gradient \citep[Pointer-PG]{pointer_haeyeon}, and AM policy gradient \citep[AM-PG]{am_hyunwook}.
    \item \textbf{Online Contextual Pretraining.} Existing online contextual DRL methods for DPP, which amortize the search process by pretraining are reported: pointer network with contextual RL \citep[Pointer-CRL]{pointer_haeyeon} and \citep[AM-CRL]{hyunwook_tmtt}.
    \item \textbf{Offline Contextual Pretraining.} Imitation learning-based contextual methods for DPP which amortize the search process with the expert dataset are reported: Pointer-CIL and AM-CIL. 
\end{itemize}

\paragraph{Dataset and Simulator}

For the DPP score simulator, a chip-package hierarchical PDN is used. The PDN model is represented as a $ N_{row} \times N_{col} = 10 \times 10$ grid over 201 frequency points linearly distributed between 200MHz and 20GHz, which gives $100 \times 100 \times 201 \approx 2M$ impedances to be evaluated per each task; simulation intensive. 

We use $N$ expert data, collected by a genetic algorithm (GA) with a specific number of simulations $M=100$ per each. For the test dataset of performance evaluation, 100 PDN cases are used. See \cref{append:pdn-decap-model} and \cref{append:DPP-generation} for detailed data construction. 

\paragraph{Implementation}
For DevFormer, we use encoder layers of $L=3$ and 128 hidden dimensions of MHA, and 512 hidden dimensions for feed-forward. See \cref{append: hyperparameter} for a detailed setup of training hyperparameters. For the implementation of baseline, methods see \cref{append: ML-baselines} and \cref{append: meta-baselines} for details.

\begin{table}[t]
\begin{center}
\caption{Performance evaluation results. We report the number of shots and average score of 100 test problems. Each method was sweeped for 5 different seeds and the average score and S.D. are reported.}

\label{Table01}\scalebox{0.9}{
\begin{tabular}{lcc}
\toprule 
Method  & Num. Shot ($M$) $\downarrow$ & Score $\uparrow$ \\
\midrule
\multicolumn{3}{c}{\textit{Online Test Time Search}}\\
\midrule
GA (\textit{expert})  & 100 & 12.56 $\pm$ 0.017\\
RS & 10,000 & 12.70 $\pm$ 0.000 \\
\midrule
\multicolumn{3}{c}{\textit{Online Test Time Adaptation}}\\
\midrule
Pointer-PG & 10,000 & 9.66 $\pm$ 0.206\\
AM-PG  & 10,000 & 9.63 $\pm$ 0.587\\
CNN-DQN  & 10,000 & 9.79 $\pm$ 0.267\\
CNN-DDQN  & 10,000 & 9.63 $\pm$ 0.150\\
\midrule
\multicolumn{3}{c}{\textit{Online Contextual Pretraining \& Zero Shot Inference}}\\
\midrule
Pointer-CRL & Zero Shot & 9.59 $\pm$ 0.232\\
AM-CRL  & Zero Shot & 9.56 $\pm$ 0.471\\
\midrule
\multicolumn{3}{c}{\textit{Offline Contextual Pretraining \& Zero Shot Inference}}\\
\midrule
Pointer-CIL  & Zero Shot & 10.49 $\pm$ 0.119\\
AM-CIL   & Zero Shot & 11.74 $\pm$ 0.075\\
\textbf{\ourmethod{}-CSE}  & \textbf{Zero Shot} & \textbf{12.88 $\pm$ 0.003}\\
\bottomrule
\end{tabular}}
\end{center}

\end{table}

\subsection{Performance Evaluation}
\label{gen-cap}
For performance evaluation, we set $N=2000$ for the offline pretraining and $M>200000 $ for online pretraining shots. The pretraining methods are evaluated with zero-shot inference at the test time. 

As shown in \cref{Table01}, our \ourmethod{} significantly outperformed all baselines in terms of average performance score. See \cref{reward} in \cref{append:objective} for the performance metric. Online search methods generally find solutions that give a high average performance. This is due to a large number of searching iterations $M$, which incurs the same number of costly simulations. On the other hand, the contextual pretrained methods including \ourmethod{} do not require simulations to generate solutions; once trained, they only require a single simulation to measure the performance after zero shot inference. \ourmethod{} is the only learning-based method capable of finding a solution that outperforms the highly iterative online search methods even by a zero-shot inference.

The online DRL methods showed poor performance in general. When the number of costly simulations was limited, RL-based methods (AM-CRL, Pointer-CRL) showed poorer generalization capability than their IL versions (AM-CIL, Pointer-CIL) due to inefficiency in exploring over extremely large combinatorial action space of DPP. We believe that the imitation learning approach, fitting the policy with offline expert data, has greater exploration capability with the help of expert policy thus able to achieve higher performance with a limited simulation budget (see \cref{append: ML-baselines}). 

Among the CIL approaches, \ourmethod{} showed the highest performance. We believe that such higher zero-shot performance comes from both symmetricity exploitation schemes and the newly devised neural architecture: (1) expert exploitation and self-exploitation with symmetric label transformation amplify the number of data to train with and induce solution symmetricity to improve generalization capability. (2) PPE, PCN, and RCN in \ourmethod{} make the policy easily adapt to new task conditions.


\paragraph{Extrapolation over Expert Method} The \ourmethod{} is trained with the offline expert data generated by the GA outperformed the GA. That is, the \ourmethod{} policy trained with lower-quality data produces higher-quality designs. We believe this is possible because we trained a factorized form of the policy that does not predict labels in a single step but produced a solution through a serial iterative roll-out process, during which a good strategy for placing decaps can be identified. 

\begin{figure}[t]
    \centering
    \begin{subfigure}[t]{0.23\textwidth}
        \centering
        \includegraphics[width=\textwidth]{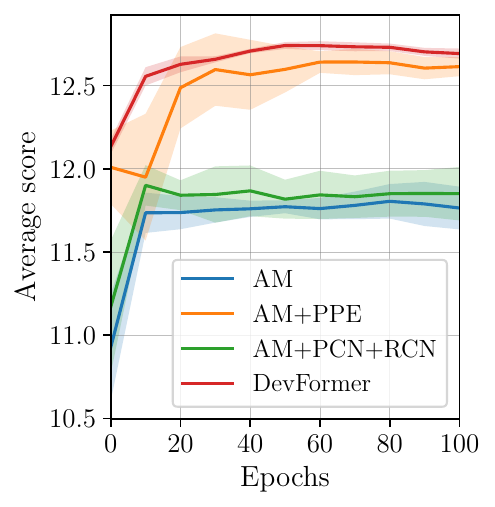}
        \vspace{-6mm}
        \caption{}
        \label{fig:ablation_b}
    \end{subfigure}
    \begin{subfigure}[t]{0.23\textwidth}
        \centering
        \includegraphics[width=\textwidth]{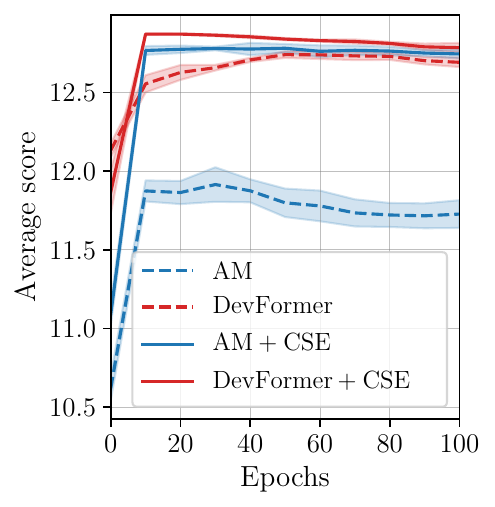}
        \vspace{-6mm}
        \caption{}
        \label{fig:component_ablation}
    \end{subfigure}
    \vspace{-1mm}
    \caption{(a) Ablation on \ourmethod{} components without CSE. Note that \ourmethod{} architecture is based on AM-CIL with newly devised PPE, PCN and RCN. AM refers to AM-CIL. (b) Ablation on AM and \ourmethod{} architecture with and without CSE learning scheme.}
    \label{fig:ablations_main_fede}

\end{figure}

\vspace{2mm}
\subsection{Ablation Study}

\paragraph{Effectiveness of \ourmethod{} Components and CSE.} We conducted ablation studies to verify the effectiveness of the proposed \ourmethod{} neural architecture and CSE training scheme. A detailed ablation on \ourmethod{} components without CSE is reported in \cref{fig:ablation_b}. The addition of PPE, PCN and RCN on top of AM-CIL baseline showed clear performance improvement. Furthermore, in \cref{fig:component_ablation}, both AM and \ourmethod{} neural architectures perform better with the presence of CSE and \ourmethod{} always outperforms the AM despite the presence of CSE. Thus, we verified that both CSE and \ourmethod{} contribute to the performance improvement.

\begin{figure}[!t]
\centering
  \begin{subfigure}[b]{0.23\textwidth}
    \centering
    \includegraphics[width=\textwidth]{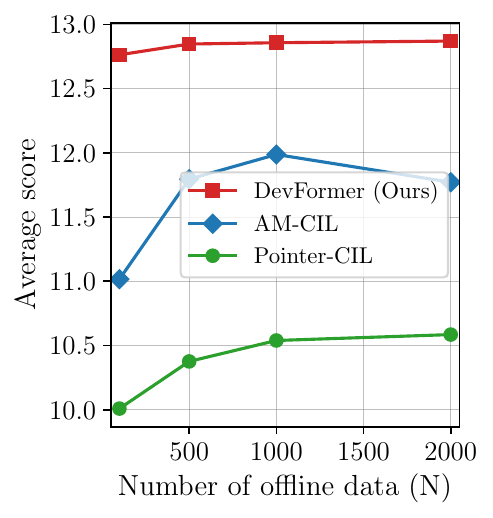}
    \vspace{-6mm}
    \caption{}
    \label{fig:N_ablation_a}
  \end{subfigure}
  \begin{subfigure}[b]{0.23\textwidth}
    \centering
    \noindent\includegraphics[width=\textwidth]{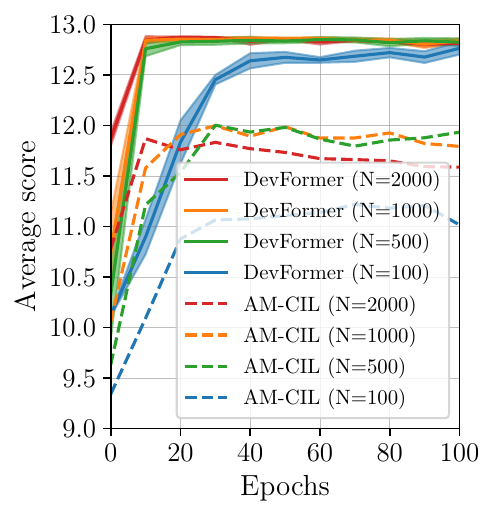}
    \vspace{-6mm}
    \caption{}
    \label{fig:N_ablation_b}
  \end{subfigure}
  \vspace{-3mm}
  \caption{Sample efficiency of \ourmethod{} in comparison to other CIL methods. (a) Performance evaluation depending on the number of offline data used for training. (b) Training convergence graph depending on the number of offline data used for training.}
  \label{fig:sample-efficiency}
\end{figure}

\begin{table}[h]
\begin{center}
\caption{ Valuation of Order Bias }
\vspace{-3mm}
\label{order-bias}
\scalebox{0.9}{
\begin{tabular}{lc}
\toprule
& $b(\pi,\boldsymbol{p}=\{\mathcal{U}_{\mathcal{X}},\pi,\mathcal{U}_{T_{AP}}\}$) \\
\midrule
AM-CIL & 8.70$\times 10^{-21}$ \\
\ourmethod{}& \boldsymbol{$1.25\times 10^{-28}$}  \\
\bottomrule
\end{tabular}}
\end{center}
\vspace{-3mm}
\end{table}

\paragraph{Order Bias Measurement} To empirically prove that our CSE induces AP-symmetricity, we measured $b(\pi,\boldsymbol{p}=\{\mathcal{U}_{\mathcal{X}},\pi,\mathcal{U}_{T_{AP}}\}$ for sample width $100$ and took the average value. As shown in \cref{order-bias}, our CSE significantly reduced the order bias defined in \cref{def:order-bias}, verifying that CSE successfully induced the AP-symmetricity.

\subsection{Sample Efficiency Evaluation for Offline Dataset}
\label{data-efficiency}

We investigated how the number $N$ of offline data generated by the GA affects the performance of \ourmethod{}, compared to AM-CIL and Pointer-CIL. As shown in \cref{fig:sample-efficiency}, \ourmethod{} outperforms the baselines in all $N$ variation; \ourmethod{} trained with $N=100$ even outperformed the baselines trained with $N=2000$.

\begin{table}[h]
\begin{center}
\caption{ Scalability evaluations on larger power distribution network (PDN) scale and a varying number of deep $K$. The $\text{scale} \times \text{scale}$ indicates the size of input grids for PDN. $K$ refers to the number of decap placed on the target PDN. Scores in bold indicate the best scores. A lower $K$ with a higher score value indicates a Pareto score. }
\vspace{-2mm}
\label{scalability}\scalebox{0.9}{
\begin{tabular}{ccccc}
\toprule
\multicolumn{2}{c}{Scale Variables}& \multicolumn{3}{c}{Methods}  \\
\cmidrule[0.5pt](lr{0.2em}){1-2} \cmidrule[0.5pt](lr{0.2em}){3-5}
PDN&{$K$} & GA & AM-CIL & Ours \\
\specialrule{0.5pt}{0pt}{4pt}
\begin{tabular}{c}
\multirow{5}{*}{10$\times$10}
\end{tabular}
&12& 11.77&10.22&\textbf{12.23}  \\
&16  & 12.25 & 11.13 &\textbf{12.60}  \\
&20 & 12.53 & 11.71 & \textbf{12.81}   \\
&24 & 12.79 & 12.20 & \textbf{12.95} \\
&30 & 13.02 & 12.62 & \textbf{13.11}  \\
\midrule
\begin{tabular}{c}\multirow{2}{*}{15$\times$15}\end{tabular}&20&7.61 & {6.23} & \textbf{8.47}\\
&40&7.69 & {7.75} & \textbf{8.54}\\
\bottomrule
\end{tabular}}
\end{center}
\vspace{-4mm}
\end{table}

\subsection {Zero shot Generalization to various tasks}
\label{section: scalability}
To verify zero shot capability on various scales of tasks, learning-based DPP methods were pretrained for a fixed scale PDN ($10 \times 10$) and a fixed number of decaps, $K=20$. Then, the pretrained models are asked to place decaps of varying $K$ on ($10 \times 10$) PDN and a larger ($15 \times 15$) PDN without additional training (i.e., zero-shot). As shown in \cref{scalability}, our \ourmethod{} outperformed GA and AM-CIL for all scales. Furthermore, \ourmethod{} achieved greater performance with fewer decaps. Reducing the number of decaps has a significant industrial impact; as hardware devices are mass-produced, reducing a single decap saves enormous fabrication costs.

\begin{figure}[!t]
    \centering
    \begin{subfigure}[b]{0.22\textwidth}
    \centering
    \includegraphics[width=\textwidth]{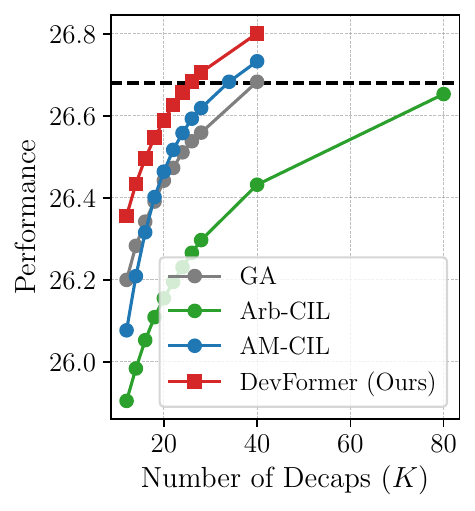}
    \vspace{-6mm}
    \caption{}
    \label{fig:hbm_decap_reduction}
    \end{subfigure}
    \begin{subfigure}[b]{0.22\textwidth}
    \centering
    \noindent\includegraphics[width=\textwidth]{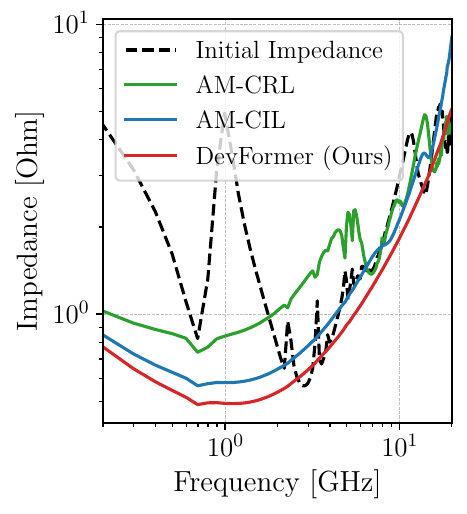}
    \vspace{-6mm}
    \caption{}
    \label{fig:impedance}
    \end{subfigure}
    \vspace{-3mm}
    \caption{(a) Performance comparison with the number of decap variation on HBM PDN. (b) Magnitude of resulting impedance suppression over wide frequency range after decap placement.}
\end{figure}

\begin{figure}[t]
\vspace{2mm}
\centering
\includegraphics[width=0.46\textwidth]{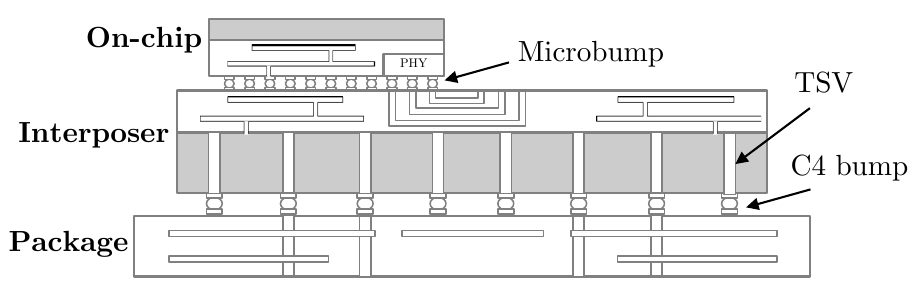}
\vspace{-1mm}
\caption{Structure of a three-layer HBM PDN model.}
\label{fig:hbm_structure}

\end{figure}


\subsection{Application on Real-world Hardware}

\label{praticality}

To verify the practical applicability of the \ourmethod{}, we applied the proposed \ourmethod{} to a real-world hardware application, the high bandwidth memory (HBM), which is an interposer-based 2.5D IC. As shown in \cref{fig:hbm_structure}, the hierarchical PDN model of HBM is composed of $(40 \times 40)$ package PDN, $(40 \times 60)$ interposer PDN and $(15 \times 20)$ on-chip PDN, each layer connected by TSV + C4 bumps and microbumps. 

For performance evaluation, we compared \ourmethod{} to GA, AM-CIL, and Pointer-CIL on placing a varying number of decaps, $K$, for 100 unseen test cases. The pretrained solvers with $K=20$ were used for \ourmethod{}, AM-CIL, and Pointer-CIL without additional training.

The results in \cref{fig:hbm_decap_reduction} demonstrate that \ourmethod{} achieves higher performance with significantly fewer decaps than other methods. The performance score of $26.68$, which was attained with $40$ decaps by the GA and $34$ decaps by AM-CIL method, was achieved with only $26$ decaps by \ourmethod{} in a zero-shot setting. The pointer-CIL method could not achieve the same performance score even with $80$ decaps. 

The resulting impedance over a wide frequency range after decap placement by each method is illustrated in \cref{fig:impedance}. It shows that decap placement by \ourmethod{} suppressed the impedance the most, leading to greater power integrity and objective value. 

We also conducted a power noise analysis on a test case (see \cref{appendix:power_noise}). \ourmethod{} was able to reduce the power noise by $94.2\%$ using only $26$ decaps, while GA reduced by $93.5\%$ using 40 decaps. As hardware devices are mass-produced, reducing even a single decap can greatly reduce production costs \cite{decap_cost}. With a more than $30\%$ reduction in the number of decaps compared to the best-performing baseline, our \ourmethod{} can significantly contribute to the industry.

\subsection{Application to Other Offline Contextual Designs}

\begin{figure}[t]
\centering
\includegraphics[width=0.45\textwidth]{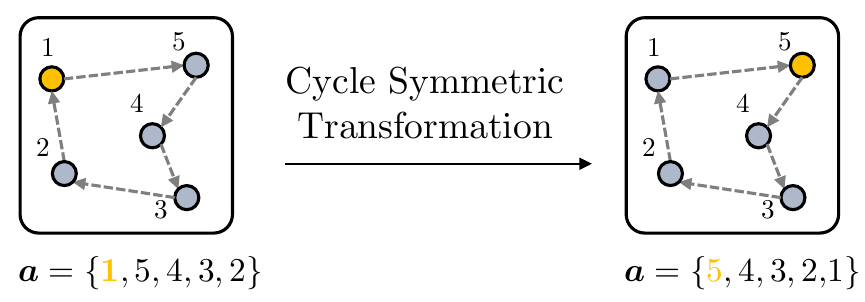}
\caption{Example of cycle symmetricity in the traveling salesman problem (TSP).}
\label{fig:tsp_sym}
\end{figure}

\vspace{-1mm}

\begin{figure}[t]
    \centering

    \begin{subfigure}[t]{0.23\textwidth}
        \centering
        \includegraphics[width=\textwidth]{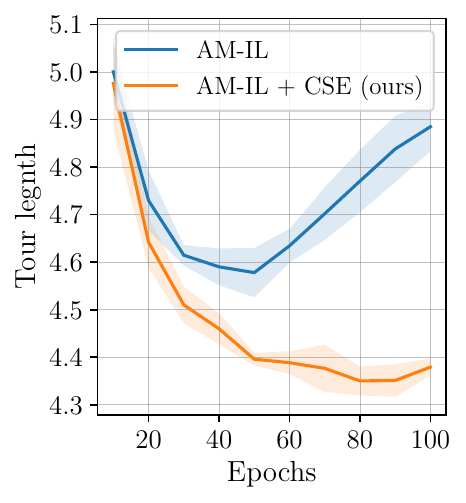}
        \vspace{-6mm}
        \caption{TSP ($N=20$)}
        \label{fig:tsp20}
    \end{subfigure}
    \begin{subfigure}[t]{0.24\textwidth}
        \centering

        \includegraphics[width=\textwidth]{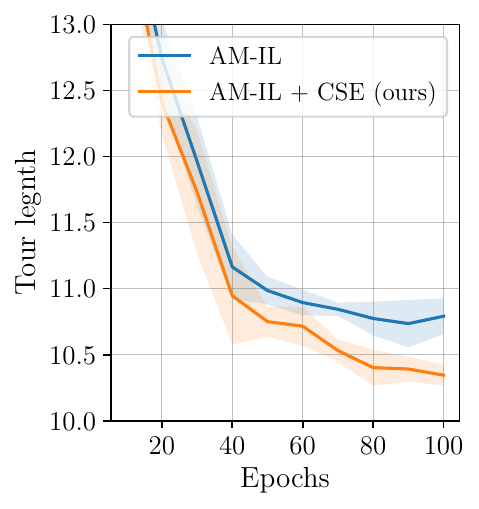}
        \vspace{-6mm}
        \caption{TSP ($N=100$)}
        \label{fig:tsp100}
    \end{subfigure}
    \vspace{-1mm}
    \caption{CSE application on TSP cyclic symmetricity. AM-IL stands for the imitation learning trained AM with sparse expert labels. The CSE greatly improves the performance of AM-IL as TSP ($N=20$) and ($N=100$).   }
    \label{fig:off-tsp}
\end{figure}

Our CSE learning scheme induces AP-symmetricity in the \ourmethod{} neural architecture and can also be applied to other architectures for offline contextual design. We tested its versatility by implementing it on the AM transformer model \citep{kool_attention} to solve the traveling salesman problem (TSP), a combinatorial optimization problem whose objective is to find the optimal tour sequence that visits all cities with minimum tour length. CSE is used to enforce cyclic symmetricity in the TSP solutions by considering the $N$ (the number of cities) symmetric solutions that can be obtained by permuting the initial cities, where the tour length is invariant (see \cref{fig:tsp_sym}).

We conducted a benchmark using a TSP problem with sparse expert data, using only 100 labels derived from the TSP Concorde \citep{concorde} solver. As shown in \cref{fig:off-tsp}, CSE significantly improves the performance of AM on 100 randomly generated synthetic test data. This indicates that our newly devised CSE can be further applied to various domains that require solution symmetricity, e.g., molecular generation, considering the symmetric nature of the molecular graph \citep{bengio2021flow}.

\vspace{4mm}
\section{Related Works}

\textbf{Symmetricity Learning in Solution Space.} Several studies have leveraged symmetricity in solution space. POMO \cite{POMO} proposed a reinforcement learning scheme that leverages symmetricity in TSP, using the cyclic property that identical solutions can be expressed as multiple permutations of initially visited nodes. GFlowNet \cite{bengio2021flow} uses a generative flow to train policy distributions proportional to reward distributions and applies a directed acyclic graph (DAG) instead of a classical tree structure to induce symmetricity. Our method is similar to POMO in that it leverages symmetricity in the solution space through regularization, but it applies this approach to imitation learning.

\textbf{New Transformers.} There have been numerous efforts to adapt the transformer architecture \citep{transformer} to new domains beyond natural language processing (NLP) \citep{gpt3}. Examples include the vision transformer (ViT) \citep{ViT}, graph transformer (Graphormer) \citep{graphormer}, attention model for combinatorial optimization (AM) \citep{kool_attention}, and decision transformer for reinforcement learning (DT) \citep{chen2021decision}. These efforts aim to incorporate domain-specific knowledge and remove unnecessary priors from the original transformer architecture, which was primarily designed for sequential data and language modeling. Our proposed method is a novel transformer for device optimization in hardware and is part of this ongoing research trend.

\section{Conclusion}

In this paper, we presented \ourmethod{}, a novel transformer model for solving contextual offline hardware design problems, and CSE, a learning scheme that induces AP-symmetricity to the neural architecture. By incorporating strong domain-specific inductive biases, our model learns more efficiently and effectively, overcoming the limitations of traditional transformer architectures. We have demonstrated the proposed approach on a novel hardware benchmark and validated its extensibility to other combinatorial optimization problems. \ourmethod{} achieved higher performance compared to other methods while considerably reducing production costs on a real-world high bandwidth memory. Future works include exploring its applicability to other hardware design tasks and investigating its scalability to larger and more complex design problems.

\section*{Acknowledgement}

We thank Hyeonah Kim, Hyunwook Park, Subin Kim, and the anonymous reviewers for their invaluable contributions in providing insightful feedback on our manuscript.

\bibliography{bibliography}
\bibliographystyle{icml2023.bst}

\newpage
\appendix
\clearpage

\newpage
\appendix
\onecolumn

\setcounter{equation}{0}
\setcounter{figure}{0}
\setcounter{table}{0}

\makeatletter
\renewcommand{\theequation}{S\arabic{equation}}
\renewcommand{\thefigure}{S\arabic{figure}}
\renewcommand{\bibnumfmt}[1]{[S#1]}
\renewcommand{\citenumfont}[1]{S#1}

\icmltitle{Appendix for ``\papertitle''}
\section{DPP Electrical Modeling and Problem Definition}
\label{append: DPP}

This section provides electrical modeling details of PDN and decap models used for verification of \ourmethod{} in DPP. Note that these electrical models can be substituted by those of the reader's interest. There are three methods to extract PDN and decap models that are also used for objective evaluation; 3D EM simulation tool, ADS circuit simulation tool, and unit-cell segmentation method. For each method, there exists a trade-off between time complexity and accuracy. See \cref{Table05}. Out of the three methods, we used the unit-cell segmentation method for a benchmark. Simulation time was evaluated using the same PDN model on a machine equipped with a 40 threads Intel\copyright~Xeon Gold 6226R CPU and 512GB of RAM. Note that simulation time depends on the size and the structural complexity of the PDN model. 

\begin{table}[h]
\vspace{-2mm}
\fontsize{9}{9}\selectfont
\begin{center}
\caption{Time Taken for an Objective Evaluation of a PDN model described in \cref{append:pdn-decap-model}}
\label{Table05}
\vspace{1mm}
\begin{tabular}{lc}
\specialrule{1.0pt}{0pt}{4pt}
Simulation Method & Time Taken\\
\specialrule{1.0pt}{0pt}{4pt}
EM Simulation Tool  & $\approx$10 hours \\
ADS Circuit Simulation Tool & 23.58 sec\\
\specialrule{1.0pt}{0pt}{4pt}
\end{tabular}
\vspace{-5mm}
\end{center}
\end{table}

\subsection{Domain Perspective Decap Placement Problem}
\label{append:domain}

\begin{figure}[t!]
    \centering
    \begin{subfigure}[b]{0.5\textwidth}
        \centering
        \includegraphics[width=\textwidth]{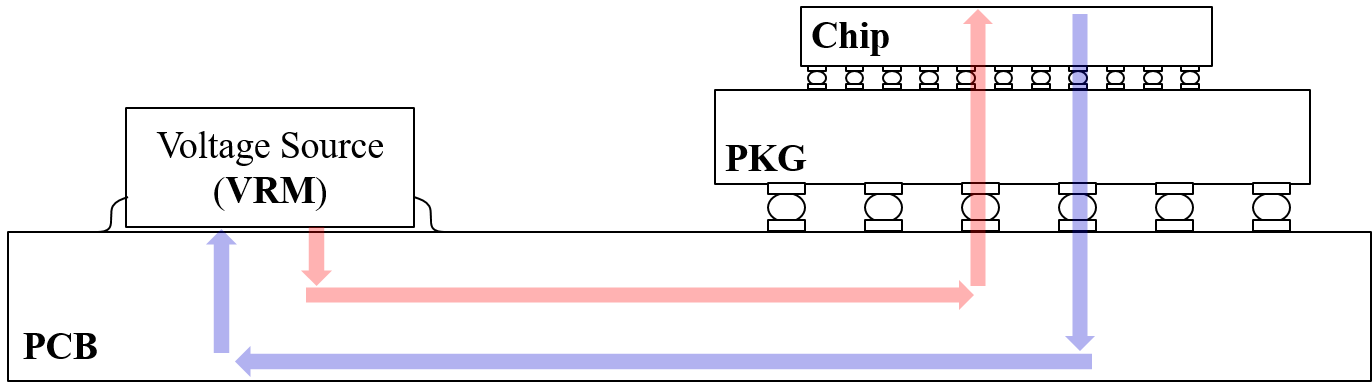}
        \caption{An example of hierarchical power distribution network (PDN).}
    \end{subfigure}

    \begin{subfigure}[b]{0.5\textwidth}
        \centering
        \includegraphics[width=\textwidth]{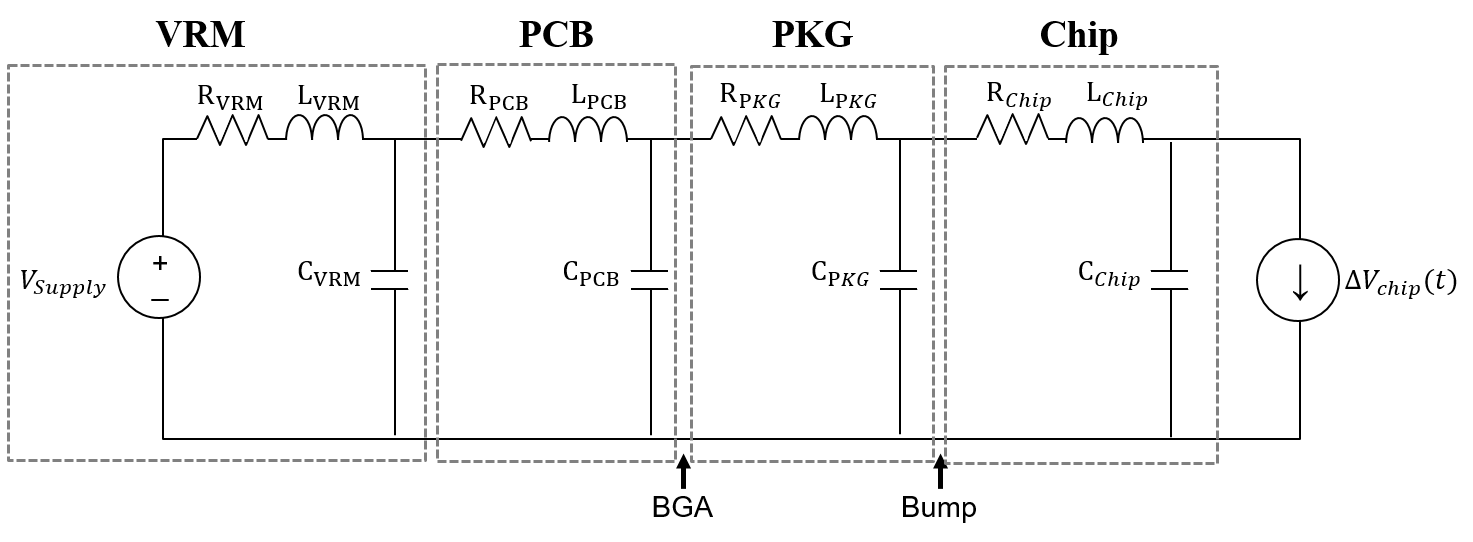}
        \caption{Electrical circuit model of the hierarchical PDN in (a).}

    \end{subfigure}

        \begin{subfigure}[b]{0.6\textwidth}
        \centering
        \includegraphics[width=\textwidth]{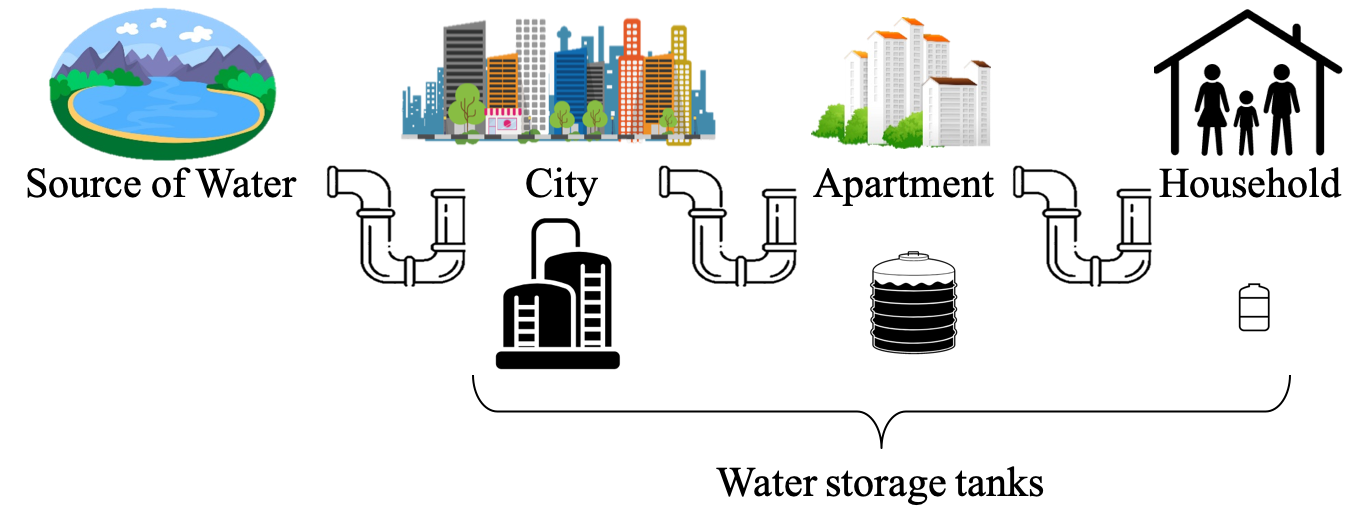}
        \caption{Water supply chain from the source to household.}
   
    \end{subfigure}
    \caption{Illustration of Hierarchical Power Distribution Network (PDN) analogous to Water Supply Chain. Similarly to how placing more water tanks can make the water supply more stable, placing more decaps can make the power supply more reliable}.
    \vspace{-6pt}
   \label{hierarchical-pdn}
\end{figure}

The development of AI has led to an increased demand for high-performance computing systems. High-performance computing systems not only require the precise design of hardware chips such as CPU, GPU, and DRAM but also require stable delivery of power to the operating integrated circuits. Power delivery has become a huge technical bottleneck of hardware devices due to the continuously decreasing supply voltage margin along with the technology shrink of the transistors \citep{voltage_margin}.

\cref{hierarchical-pdn} (a) shows the power distribution network (PDN) consisting of all the power/ground planes from the voltage source to operating chips. Power is generated in the voltage regulator module (VRM) and delivered through electrical interconnections of PCB, package and chip. Finding ways to meet the desired voltage and current from the power source to destinations along the PDN is detrimental because failure in achieving power integrity (PI) leads to various reliability problems such as incorrect switching of transistors, crosstalk from neighboring signals, and timing margin errors \citep{PI_textbook}. Decoupling capacitors (decaps) placed on the PDN allows a reliable power supply to the operating chips, thus improving the power integrity of hardware. As shown in \cref{hierarchical-pdn} (b)-(c), the role of decap is analogous to that of water storage tanks, placed along the city, apartment, and household, that can provide water uninterruptedly and reliably. As placing more water tanks can make the water supply more stable, placing more decaps can make the power supply more reliable. 
However, because adding more decaps requires more space and is costly, optimal placement of decaps is important in terms of PI and cost/space-saving.

\subsection{PDN and Decap Models for Verification}
\label{append:pdn-decap-model}

\textbf{Unit-Cell Segmentation Method.} The segmentation method \citep{segmentation} is a simple and fast way to generate approximated electrical models. Because the analysis of the full electrical model using EM simulation is very time-consuming, we divided the full PDN model into smaller unit-cells and constructed the full PDN model using the unit-cell segmentation method. For fast simulation, we used an equation-based implementation in Python of the segmentation method, illustrated in \cref{segmentation}. 

The segmentation method was used for the generation of the PDN model consisting of a chip layer and a package layer for verification as illustrated in \cref{segmentation} (a). The segmentation method was also used for the objective evaluation of DPP. When a solution for DPP is made, decaps are placed on the corresponding ports on PDN using the segmentation method as illustrated in \cref{segmentation} (b).

\begin{figure}[h!]
    \centering
    \begin{subfigure}[b]{0.35\textwidth}
        \centering
        \includegraphics[width=\textwidth]{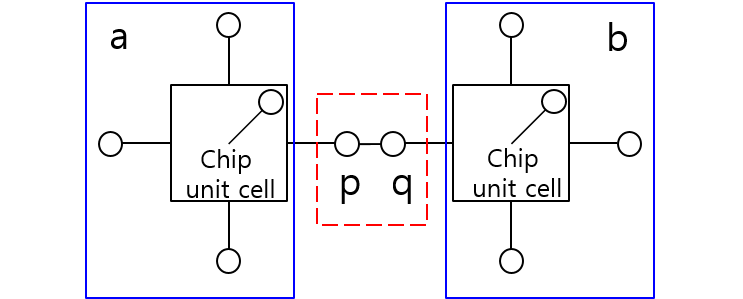}
        \caption{Generation of Chip PDN.}
    \end{subfigure}

    \begin{subfigure}[b]{0.35\textwidth}
        \centering
        \includegraphics[width=\textwidth]{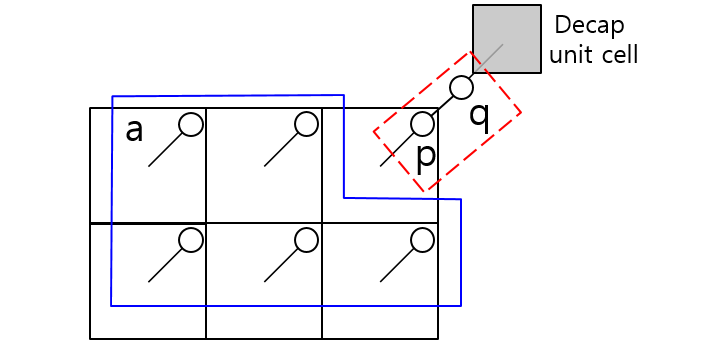}
        \caption{Decap Placement.}

    \end{subfigure}

        \begin{subfigure}[b]{0.6\textwidth}
        \centering
        \includegraphics[width=\textwidth]{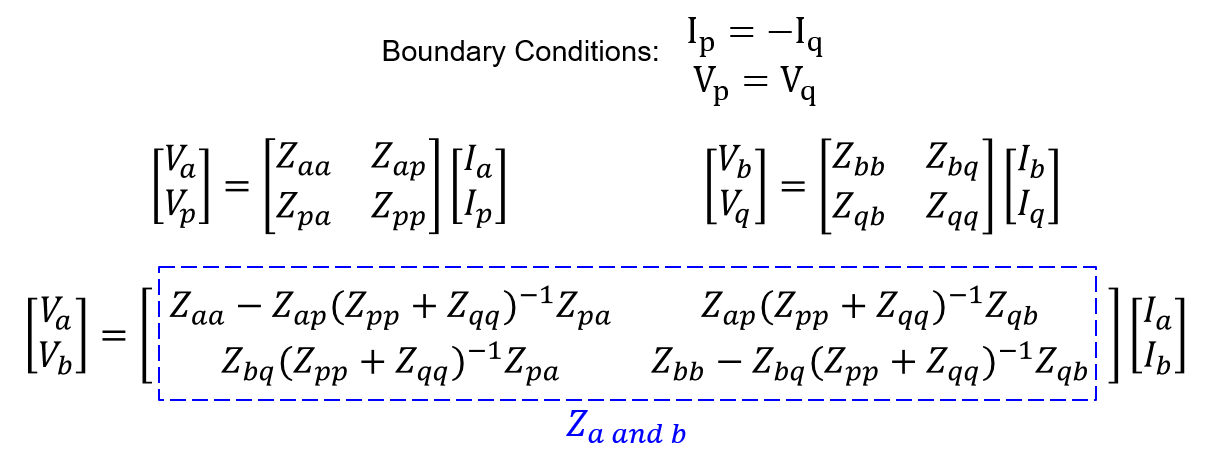}
        \caption{Segmentation Method.}
   
    \end{subfigure}
    \caption{Segmentation Method Implemented for PDN Generation and Decap Placement on PDN.}
    \vspace{-6pt}
   \label{segmentation}
\end{figure}

The PDN model we used for verification has a two-layer structure; a package layer at the bottom and a chip layer on top of it as illustrated in \cref{pdn}. The PDN was modeled through the unit-cell segmentation method. The package layer was composed of $40\times40$ package unit-cells and the chip layer was composed of $10\times10$ (i.e, $N_{row} \times N_{col}$) chip unit-cells. Because the DPP benchmark places MOS-type decaps, which are placed on the chip, ports are only available on the chip. Thus, we extracted $10 \times 10$ ports information from the chip layer. See \cref{input_output} (a), illustrating the chip PDN divided into $10\times10$ units and each unit-cell numbered.

\begin{figure}[h!]
    \centering
    \begin{subfigure}[b]{0.32\textwidth}
        \centering
        \includegraphics[width=\textwidth]{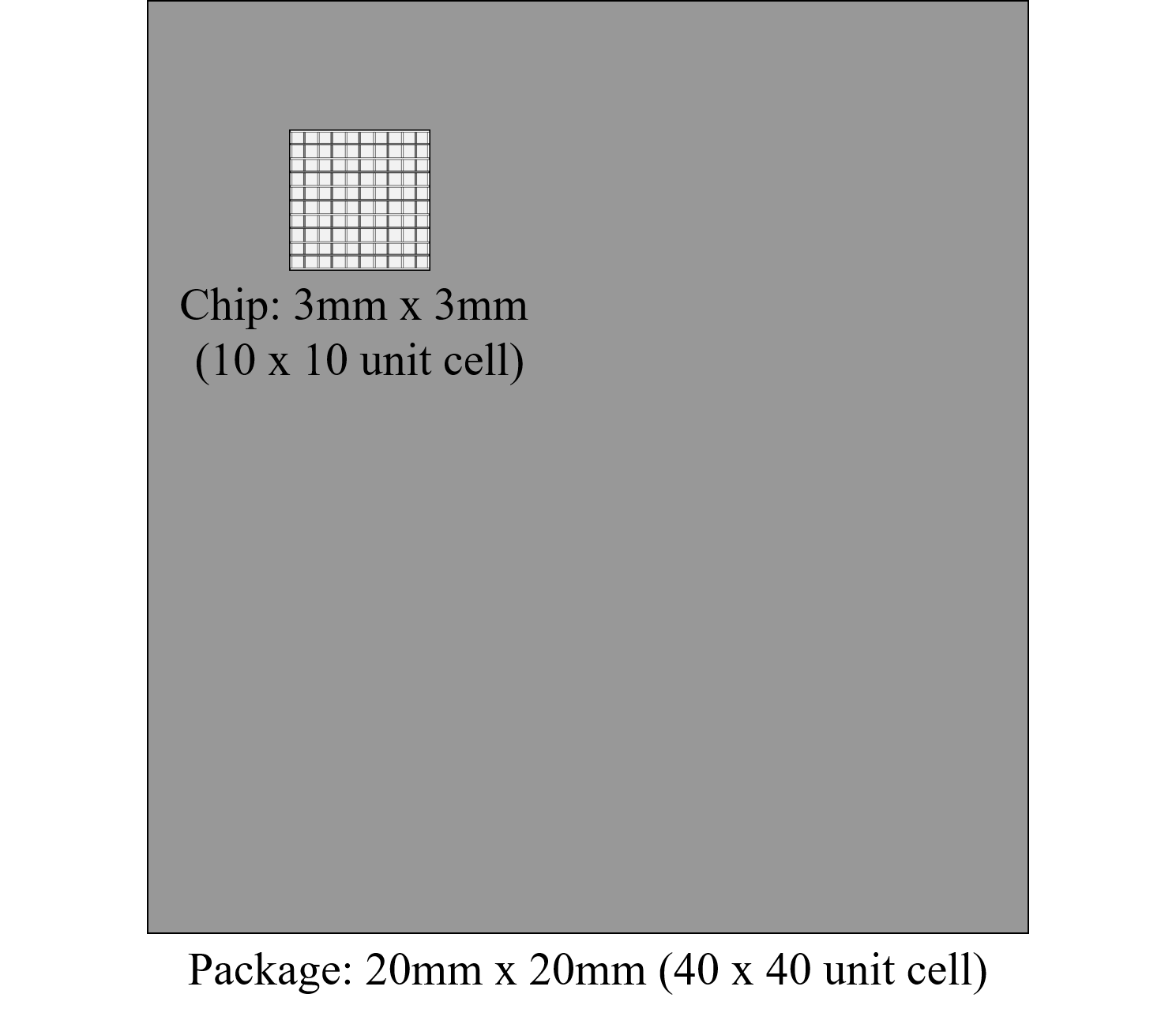}
        \caption{Top-View of PDN model.}
    \end{subfigure}
    \begin{subfigure}[b]{0.52\textwidth}
        \centering
        \includegraphics[width=\textwidth]{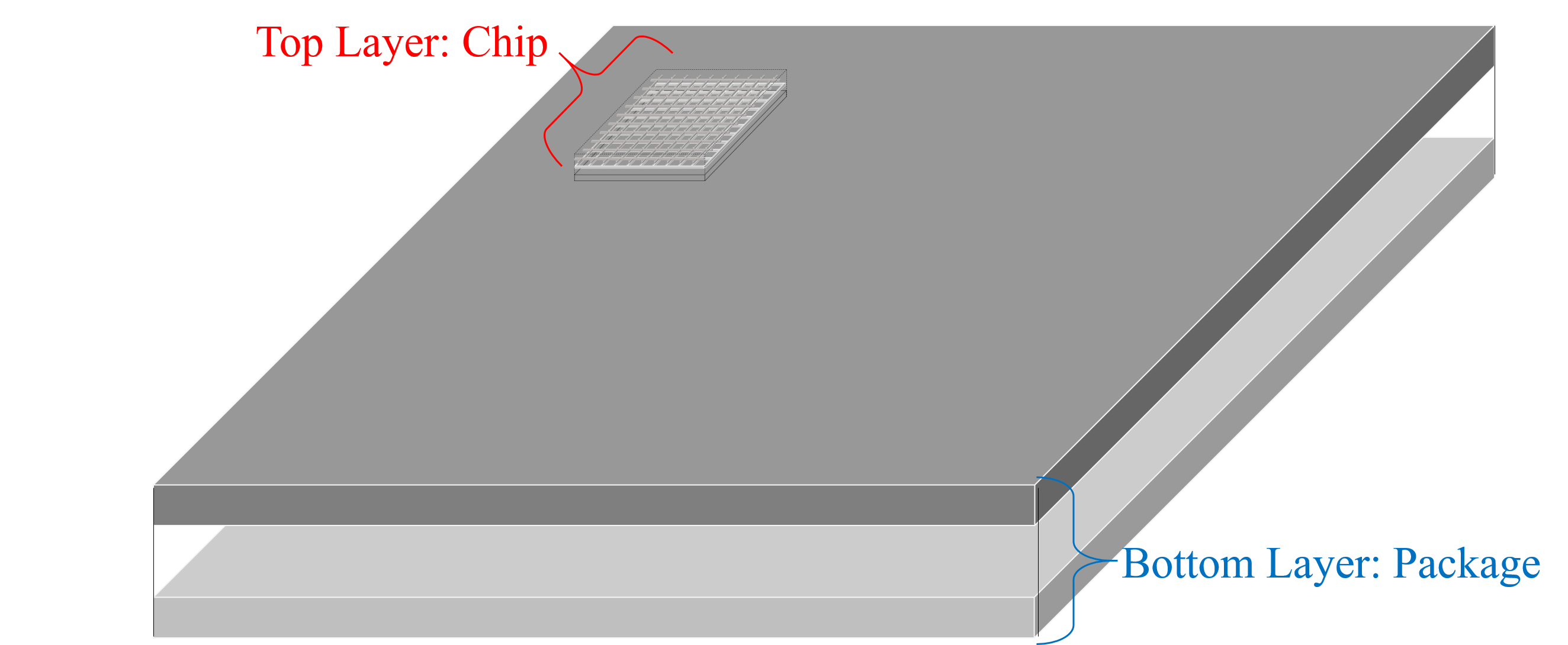}
        \caption{Side-View of PDN model.}
    \end{subfigure}
    \caption{Top-view and Side-view of PDN Model used for Verification}
   \label{pdn}
\end{figure}

The electrical models of package and chip unit-cells that are used to build the PDN model for verification are described in \cref{unitcell}. The chip layer is composed of $10\times10$ unit-cells, and the package layer is composed of $40\times40$ unit-cells using the segmentation method. The corresponding values of electrical parameters are listed in \cref{Table06}.

\begin{figure}[h!]
    \centering
    \begin{subfigure}[b]{1\textwidth}
        \centering
        \includegraphics[width=\textwidth]{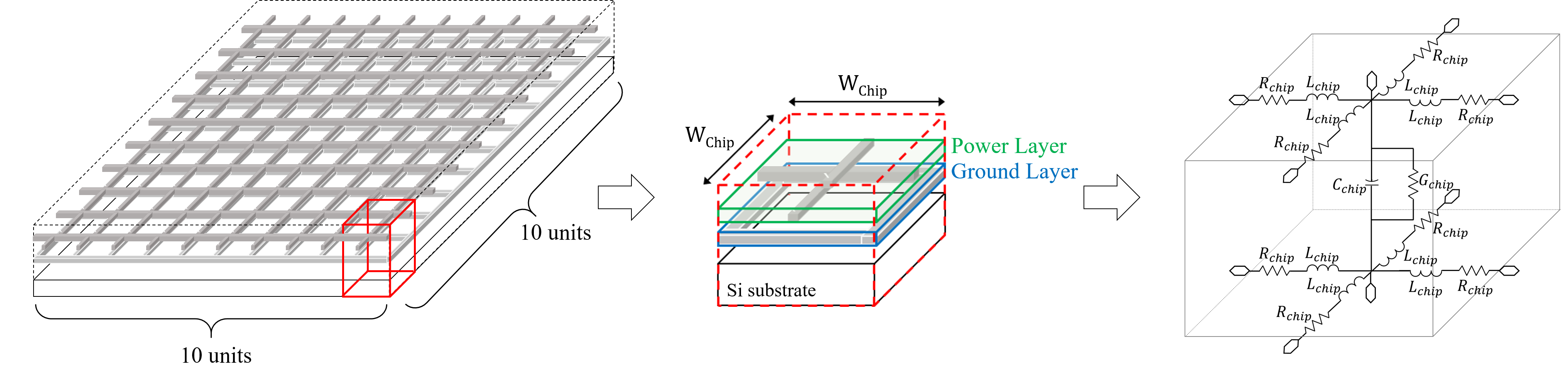}
        \caption{Balanced Transmission Line Model of Chip Unit-Cell.}
    \end{subfigure}
    \begin{subfigure}[b]{1\textwidth}
        \centering
        \includegraphics[width=\textwidth]{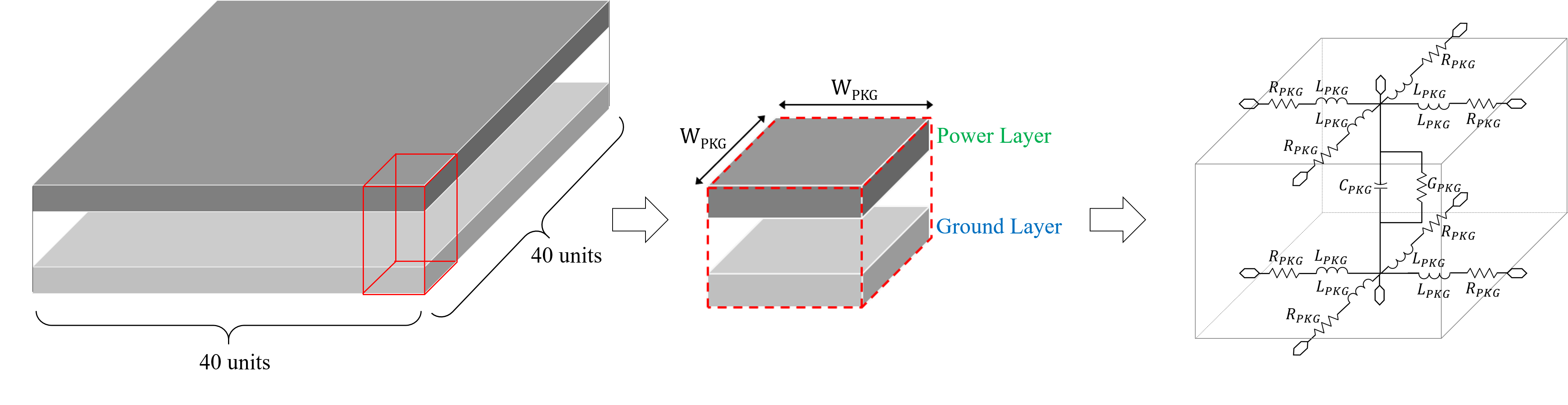}
        \caption{Balanced Transmission Line Model of Package Unit-Cell.}
    \end{subfigure}
    \caption{Electrical Modeling of Chip and Package Unit-Cells for PDN Model generation.}
   \label{unitcell}
\end{figure}

\begin{table}[h!]
\fontsize{9}{9}\selectfont
\begin{center}
\caption{{Width and Electrical Parameters for Chip and Package Unit-Cells used for Verification.}}\label{Table06}
\vspace{1mm}
\begin{tabular}{lccccc}
\specialrule{1.0pt}{0pt}{4pt}
Unit-Cell Model & W & R & L & G & C\\
\specialrule{1.0pt}{0pt}{4pt}
Chip & 300$\mu\mathrm{m}$ & 0.26 $\Omega$ & 22$\mathrm{pH}$ & 1.2{$\mathrm{m}\mathrm{S}$} & 0.77$\mathrm{pF}$\\
Package & 0.5$\mathrm{mm}$ & 0.093 $\Omega$ & 0.25$\mathrm{nH}$ & 5.4{$\mu\mathrm{S}$} & 0.045$\mathrm{pF}$\\
\specialrule{1.0pt}{0pt}{4pt}
\end{tabular}
\end{center}
\end{table}

\begin{figure}[t]
\centerline{\includegraphics[width=0.23\textwidth]{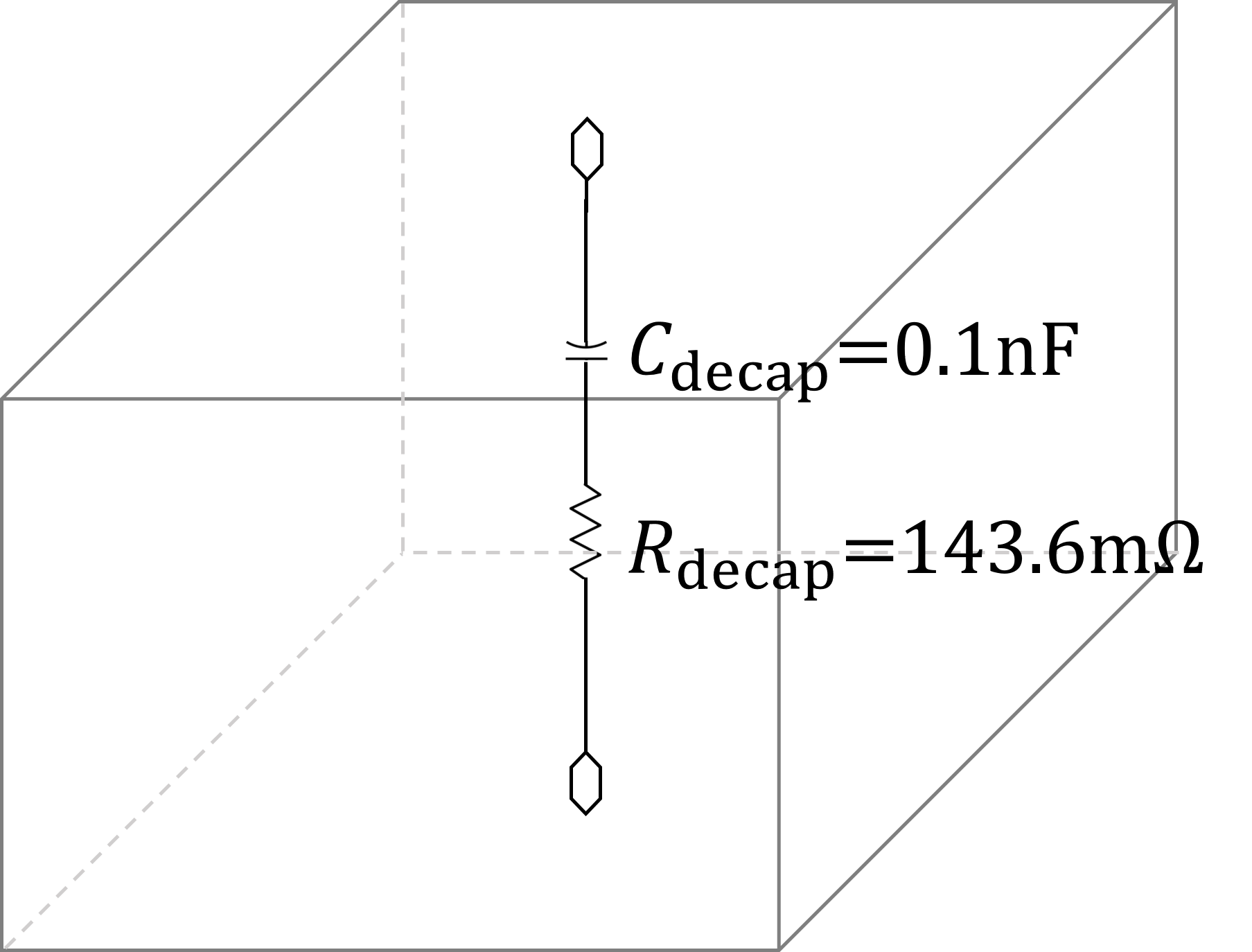}}
\caption{Decap Unit-Cell with the Electrical Parameters used for Verification.}
\label{decap}
\end{figure}

We implemented MOS type decap for verification. The Decap model and its electrical parameters are shown in \cref{decap}. As mentioned in \cref{segmentation} (b), the solution to DPP is evaluated using the segmentation method. 

Note that these electrical parameters and PDN structures were used as a benchmark. For practical use of \ourmethod{}, these PDN and decap models can be substituted by those of the reader's interest.

\subsection{Input Problem PDN and Output Decap Placement Data Structure}
\label{append:data-structure}

\begin{figure}[h!]
    \centering
    \begin{subfigure}[b]{0.45\textwidth}
        \centering
        \includegraphics[width=\textwidth]{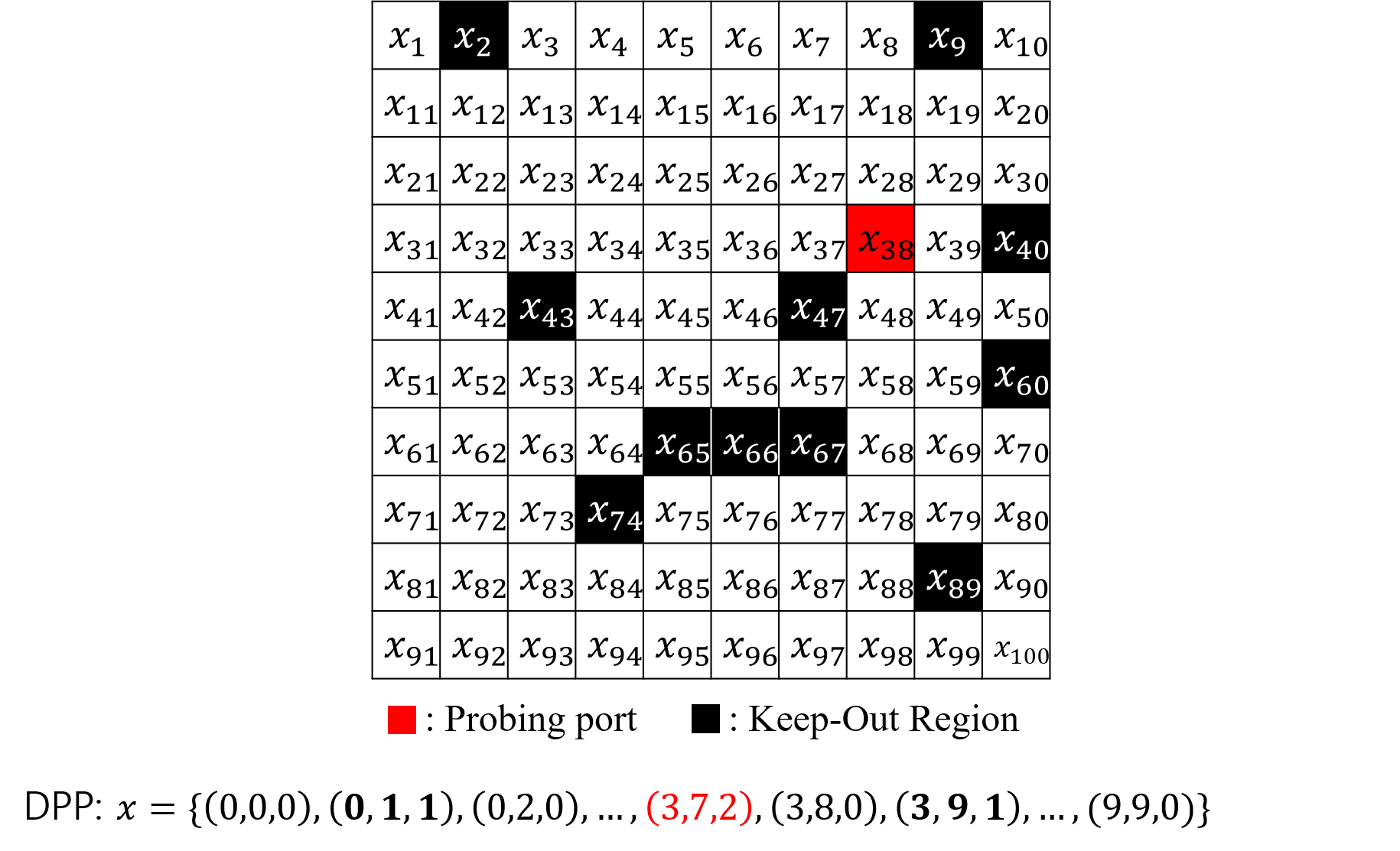}
        \caption{Input Problem PDN.}
    \end{subfigure}
    \begin{subfigure}[b]{0.45\textwidth}
        \centering
        \includegraphics[width=\textwidth]{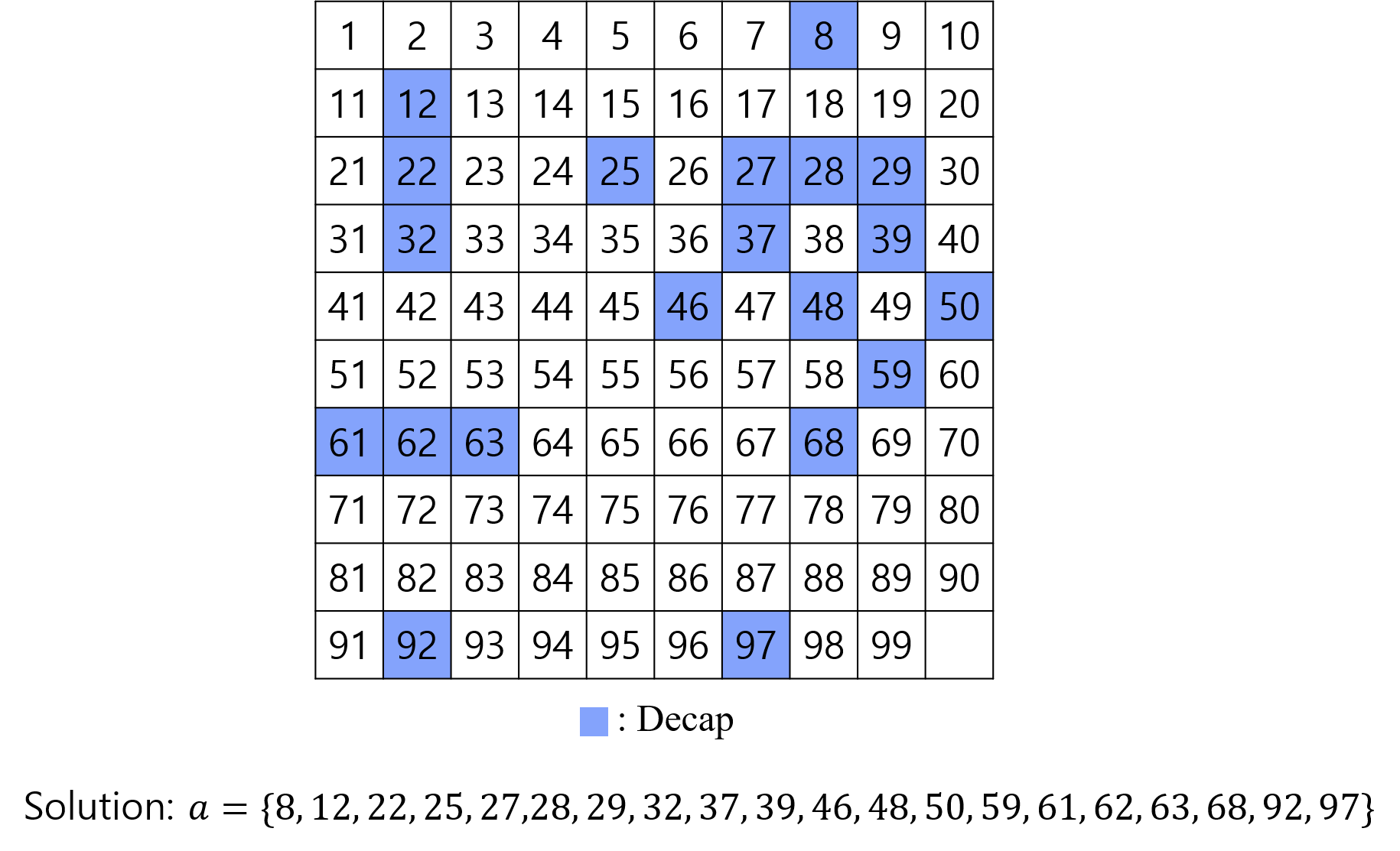}
        \caption{Output Decap Placement Solution.}
    \end{subfigure}
    \caption{Illustration of how the DPP problem with specific conditions is given as an input and decap placement solution is generated as an output.}
    \label{input_output}
\end{figure}

Each unit-cell (i.e, port) of the PDN model described in \cref{append:pdn-decap-model} is represented as a set of 3D feature vectors composed of x-coordinate, y-coordinate and port condition; 1 representing the keep-out region, 2 representing a probing port and 0 for the decap allowed ports. Total $10\times10$ (i.e, $N_{row} \times N_{col}$) 3D vectors represent the problem PDN. The solution to DPP is the placement of decaps. As illustrated in \cref{input_output} (b), the solution is given as a set of port numbers corresponding to each decap location. 

\subsection{Objective Function of DPP}
\label{append:objective}

The objective of DPP is evaluated by power integrity (PI) simulation that computes the level of impedance suppression over a specified frequency domain:
\vspace{-0.5mm}
\begin{equation}
    \mathcal{J} :=\sum_{f \in F} (Z_{initial}(f)-Z_{final}(f)) \cdot \frac{\text{1GHz}}{f}
    \label{reward}
\end{equation}
where $Z_{initial}$ and $Z_{final}$ are the initial and final impedance at the frequency $f$ before and after placing decaps, respectively. $F$ is the set of specified frequency points. The more impedance is suppressed, the better the power integrity and the higher the performance score. Remark that DPP cannot be formulated as a conventional mixed-integer linear programming (MILP)-based combinatorial optimization because PI performance can not be formulated as a closed analytical form but can only be measured or simulated. 

\subsection{Random Problem Generation of DPP}
\label{append:DPP-generation}

To randomly generate decap placement problems (DPPs) with distinct conditions for training, testing and validation, a probing index $I_{probe}$ is selected randomly from a uniform distribution of $\{1,...,N_{row}\times N_{col}$\}. Then keep-out region indices $I_{keepout}$ are randomly selected through the following two stages: the number of keep-out regions $|I_{keepout}|$ is randomly selected from a uniform distribution of $0\sim15$. Then, a set of  indices of keep-out ports $I_{keepout}$ is generated by random selection from the uniform distribution of $\{1,...,N_{row}\times N_{col}$\}. We generated 100 test problems and 100 validation problems for $10\times10$ PDN and 50 test problems and 50 validation problems for $15\times15$ PDN. We made sure the training, test, and validation problems do not overlap.

\section{Expert Label Collection}
\label{append: expert}

\begin{figure}[h!]
\centering
\centerline{\includegraphics[width=0.4\textwidth]{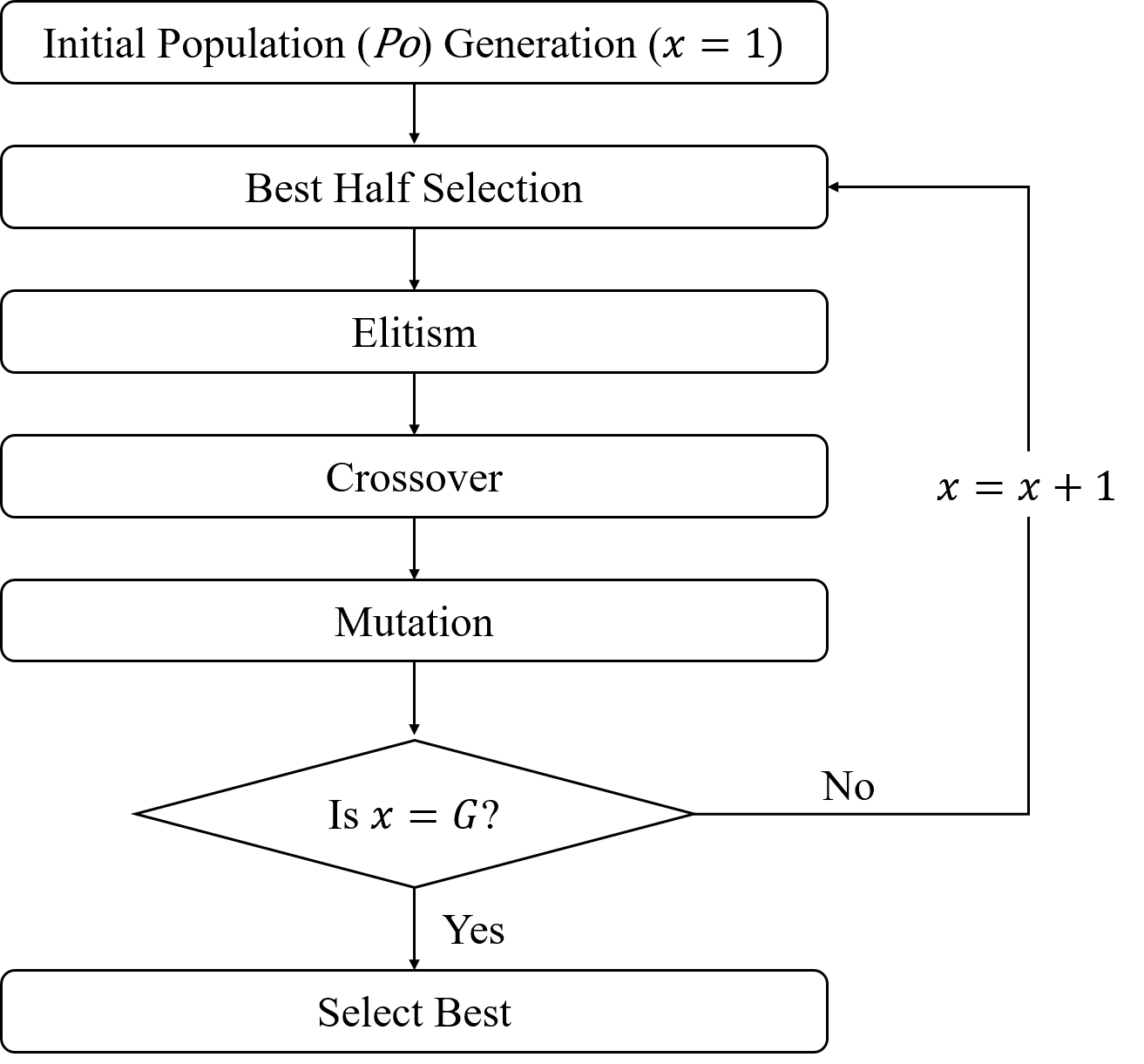}}
\caption{Process Flow of Genetic Algorithm for DPP.}
\label{ga_process}
\end{figure}

We used a genetic algorithm (GA) as the expert policy to collect expert guiding labels for imitation learning. GA is the most widely used search heuristic method for DPP \citep{ga6, ga8, ga9, ga10}. We devised our own GA for DPP, which aims to find the placement of a given number ($K$) of decaps on PDN with a probing port and 0-15 keep-out regions that best suppresses the impedance of the probing port. 

\textbf{Notations.} 
$M$ is the number of samples to undergo an objective evaluation to give the best solution. The value of $M$ is defined by the size of population $P_0$ times the number of generation $G$. $K$ refers to the number of decaps to be placed. $P_{elite}$ is the number of elite population.

\textbf{Guiding Dataset.}
To generate expert labels, guiding problems were generated in the same way the test dataset was generated. We ensured the guiding data problems do not overlap with the test dataset problems. Also, we made sure each guiding problem does not overlap. Each guiding data problem goes through the process described in \cref{ga_process} to collect the corresponding expert label.

\begin{figure}[t!]
    \centering
    \begin{subfigure}[b]{0.48\linewidth}
        \centering
        \includegraphics[width=\textwidth]{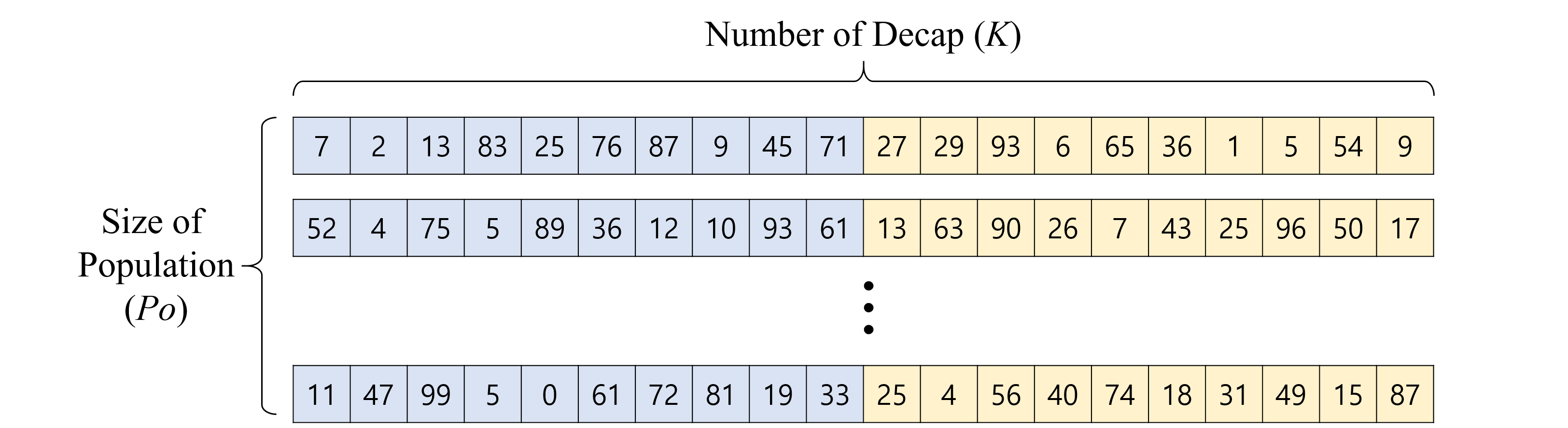}
        \caption{Initial Population Generation.}
    \end{subfigure}
    \begin{subfigure}[b]{0.48\linewidth}
        \centering
        \includegraphics[width=\textwidth]{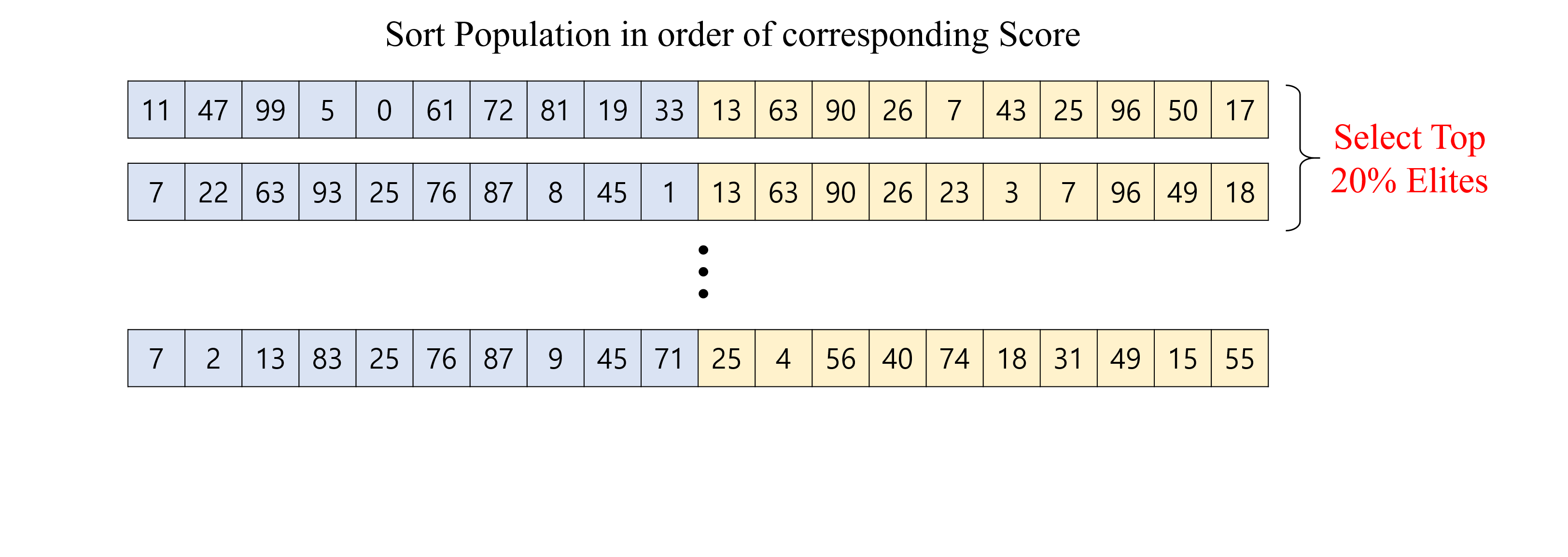}
        \caption{Elitism.}
    \end{subfigure}
    \begin{subfigure}[b]{0.48\linewidth}
        \centering
        \includegraphics[width=\textwidth]{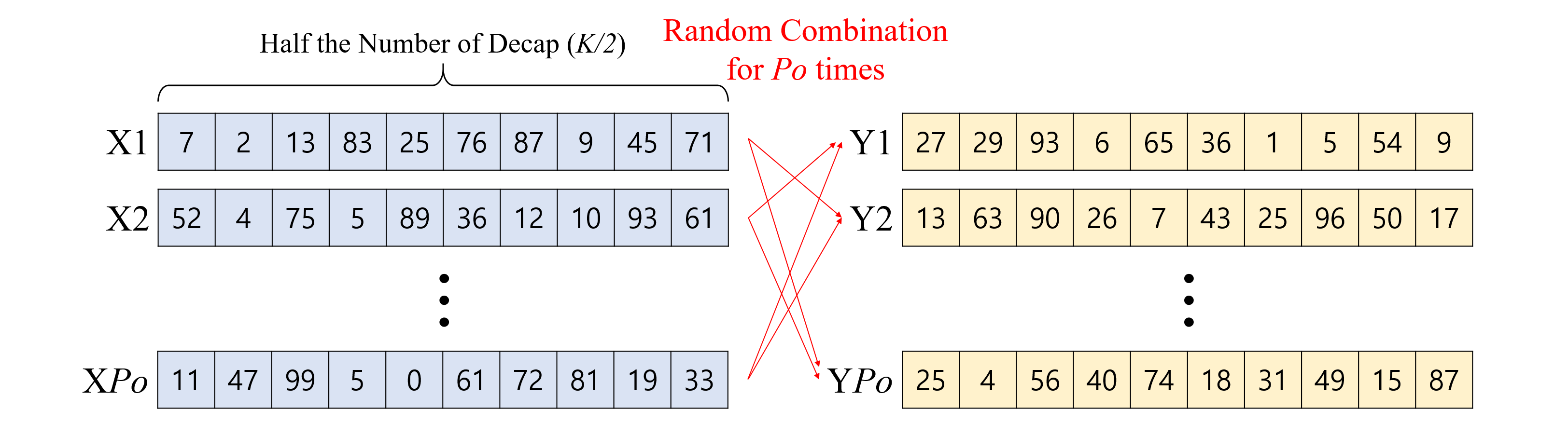}
        \caption{Crossover.}
    \end{subfigure}
    \begin{subfigure}[b]{0.48\linewidth}
        \centering
        \includegraphics[width=\textwidth]{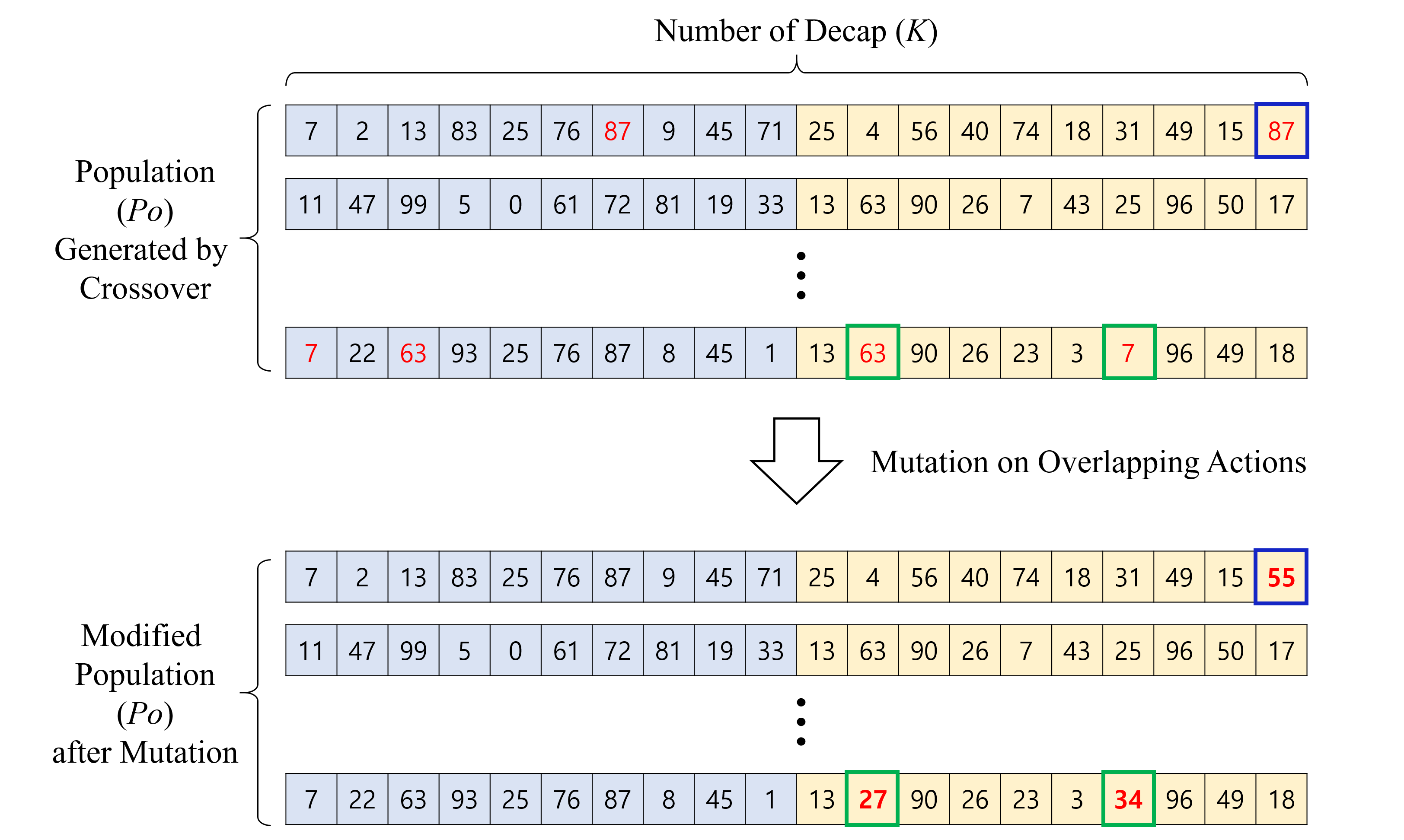}
        \caption{Mutation.}
    \end{subfigure}

    \caption{Illustration of each GA Operators used for DPP Guiding Data Generation.}
    \label{ga}
\end{figure}

\textbf{Population and Generation.}
For GA $\{M=100\}$ (\textit{expert policy}), we fixed the population size as $P_0=20$ and the generation number $G=5$, which makes up the total number of samples to be $M=P_0 \times G=100$. Each solution in the initial population is generated randomly. As described in \cref{input_output} (b), each solution consists of $K$ numbers, each representing a decap location on PDN. Note that each solution consists of random numbers from 0 to 99 except numbers corresponding to probing port and keep-out region locations. 

Once the initial population is generated randomly, a new population is generated through elitism, crossover, and mutation. This whole process of generating a new population makes one generation; the generation process is iterated for $G-1$ times.

\textbf{Elitism.}
Once the initial population is formulated, the entire population undergoes objective evaluation and gets sorted in order of objective value. The size of elite population is pre-defined as $P_{elite}=4$ for GA $\{M=100\}$ (\textit{expert policy}). That means the top 4 solutions in the population become the elite population and are kept for the next generation.

\textbf{Crossover.}
Crossover is a process by which new population candidates are generated. Each solution of the current population, including the elites, is divided in half. Then, as described in \cref{ga} (c), half the solutions on the left and the other half on the right go through random crossover for $P_0$ times to generate a new population. If the elite population is available, $P_0-P_{elite}$ random crossover takes place so that the total population size becomes $P_0$, including the elite population. 

\textbf{Mutation.}
According to \cref{ga} (d), solutions with overlapping numbers may exist after the random crossover. We replace the overlapping number with a randomly generated number, and we call this mutation. 

\textbf{Select Best.}
When $G$ is reached, the final population is evaluated by the performance metric. Then, a solution with the highest objective value becomes the final guiding solution for the given DPP.

The guiding problems and corresponding solutions generated from GA are saved and used as guiding expert labels for imitation learning. 
\newpage
\section{Detailed Experimental Settings} 
\label{append: exp_settings}

This section provides detailed experimental settings for main experiments and ablation studies.

\subsection{Training Hyperparameters.}
\label{append: hyperparameter}

 There are several hyperparameters for training; we tried to fix the hyperparameters as \citet{kool_attention} did to show their frameworks' practicality. We then provided several ablation studies on each hyperparameter to analyze how each component contributes to performance improvement.
 
Training hyperparameters are set to be identical to those presented in AM for TSP \citep{kool_attention} except learning rate, unsupervised regularization rate $\lambda$, the number of expert data $N$, number of action permutation transformed data per expert data $P$ and batch size $B$. 

\vspace{-2mm}
\begin{table}[h]
\fontsize{9}{9}\selectfont
\begin{center}
\caption{{Hyperparameter setting for training model.}}
\label{hyperparameters}
\vspace{-2mm}
\begin{tabular}{lc}
\specialrule{1.0pt}{0pt}{4pt}
Hyperparameter & Value\\
\specialrule{0.5pt}{0pt}{4pt}
learning rate & $10^{-5}$ \\
$\lambda$ & $\times 10^{32}$ \\
$N$ & 2000 \\
$P$ & 4 \\
$B$ & 100 \\
\specialrule{1.0pt}{0pt}{4pt}
\end{tabular}
\end{center}
\end{table}
\vspace{-4mm}

\subsection{Implementation of ML Baselines.}
\label{append: ML-baselines}

There are two main ML baselines, Pointer-CRL \citep{pointer_haeyeon} and AM-CRL \citep{hyunwook_decap}. 

\textbf{Pointer-CRL.} Pointer-CRL is a PointerNet-based DPP solver proposed by \citet{pointer_haeyeon}. However, reproducible source code was not available. Therefore, we implemented the Pointer-CRL following the implementation of \citet{bello_pointer} \footnote{\href{https://github.com/pemami4911/neural-combinatorial-rl-pytorch}{https://github.com/pemami4911/neural-combinatorial-rl-pytorch}} and paper of \citet{pointer_haeyeon}. We set the training step $1,600$ with batch size $B=100$, which makes a total $160,000$ PI simulations.    

\textbf{Pointer-CIL.} Pointer-CIL is an imitation learning version of Pointer-CRL trained by our training data. We set $N=2000$, $B=1000$ for training Pointer-CIL. 

\textbf{AM-CRL.} AM-CRL is a AM-based DPP solver proposed by \citet{hyunwook_decap}. We reproduced AM-CRL by following implementation of \citet{kool_attention}\footnote{\href{https://github.com/wouterkool/attention-learn-to-route}{https://github.com/wouterkool/attention-learn-to-route}} and paper of \citet{hyunwook_decap}. We set the training step $2,000$ with batch size $B=100$, which makes a total of $200,000$ PI simulations. 

\textbf{AM-CIL.} AM-CIL is an imitation learning version of AM-CRL trained by our training data. For experiments in Table 1, we set $N=2000$ and $B=1000$ for training. For the ablation study, we mainly ablate $N$, when $N = 100$ we set $B=100$. Here is the training sample complexity (the number of PI simulations during training) of each ML baseline and \ourmethod{}: 

\begin{table}[h]
\fontsize{9}{9}\selectfont
\begin{center}
\caption{{Training sample complexity of pre-trained ML baselines and \ourmethod{}.}}
\label{sample_complexity}
\vspace{-2mm}
\begin{tabular}{lc}
\specialrule{1.0pt}{0pt}{4pt}
Methods & Number of PI simulations for Training \\
\specialrule{0.5pt}{0pt}{4pt}
Pointer-CRL & 200,000 \\
AM-CRL & 200,000 \\
Pointer-CIL \{$N=2000$\} & 200,000 ($N=2000$, $M=100$ from GA expert) \\
AM-CIL \{$N=2000$\} & 200,000 ($N=2000$, $M=100$ from GA expert) \\
\specialrule{0.5pt}{0pt}{4pt}
\textbf{\ourmethod{}} \{$N=100$\} (ours) & 10,000 ($N=100$, $M=100$ from GA expert) \\
\textbf{\ourmethod{}} \{$N=500$\} (ours) & 50,000 ($N=500$, $M=100$ from GA expert) \\
\textbf{\ourmethod{}} \{$N=1000$\} (ours) & 100,000 ($N=1000$, $M=100$ from GA expert) \\
\textbf{\ourmethod{}} \{$N=2000$\} (ours) & 200,000 ($N=2000$, $M=100$ from GA expert) \\
\specialrule{1.0pt}{0pt}{4pt}
\end{tabular}
\end{center}
\end{table}
\vspace{-2mm}


During the inference phase, each learned model produces a greedy solution from their policies (i.e., $M=1$) following \citep{kool_attention}.

\newpage
\subsection{Implementation of Meta-Heuristic Baselines.}
\label{append: meta-baselines}

\textbf{Genetic Algorithm (GA).}
GA $\{M=100\}$ and GA $\{M=500\}$ are implemented as baselines. For detailed procedures and operators used for GA, see \cref{append: expert}. GA $\{M=100\}$ is the expert policy used to generate expert data for imitation learning in \ourmethod{}. For GA $\{M=100\}$, the size of population, $P_0$, is 20, number of generation, $G$, is 5 and elite population, $P_{elite}$, is 4. For GA $\{M=500\}$, $P_0$ is 50, $G$ is 10 and $P_{elite}$ is 10. 

\textbf{Random Search (RS).}
The random search method generates $M$ random samples for a given problem and selects the best sample with the highest objective value.

\begin{figure}[h]
\centerline{\includegraphics[width=0.65\textwidth]{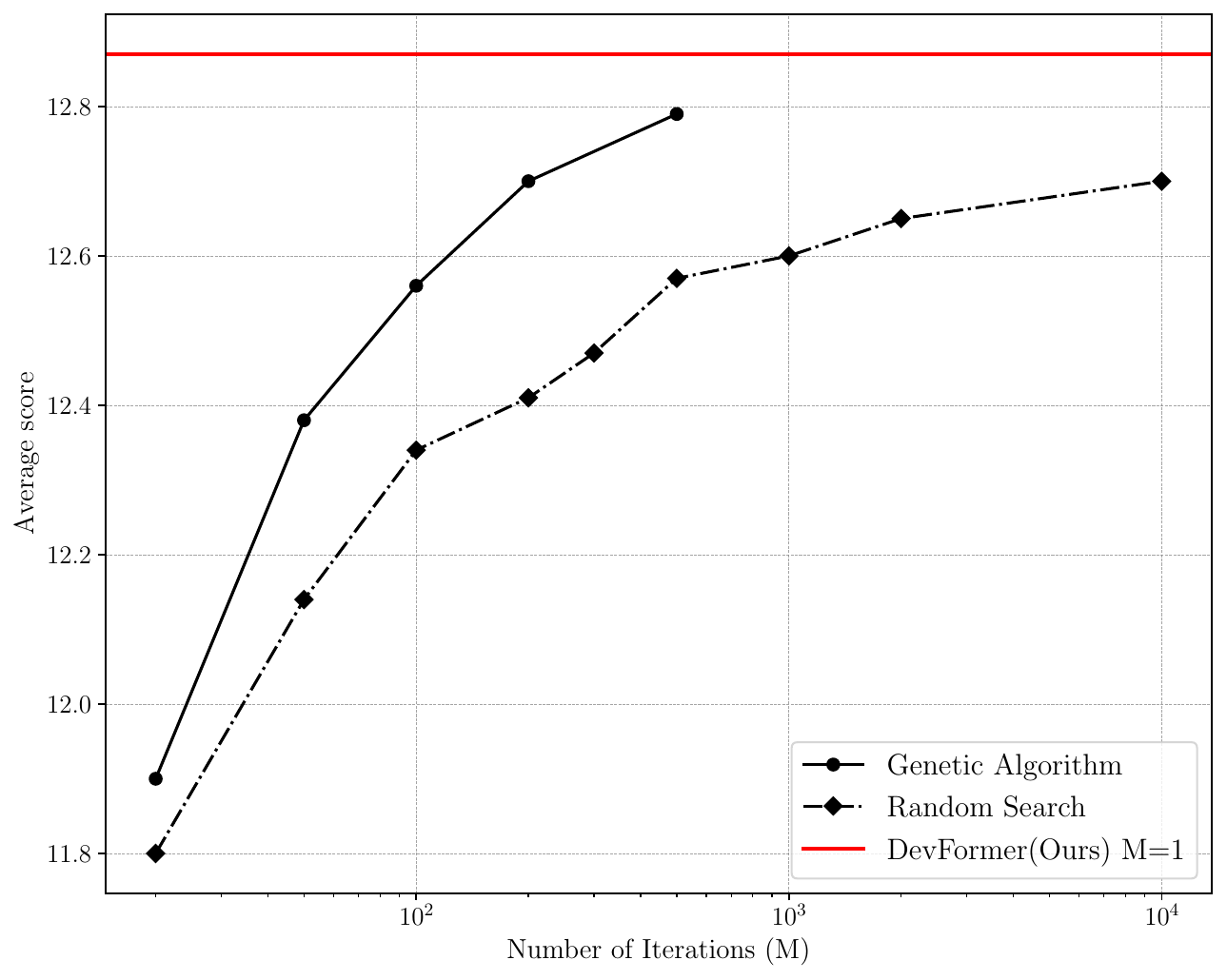}}
\caption{Performance of GA and RS with a varying number of iterations ($M$) in comparison to \ourmethod{} at $M=1$.}
\label{M_performance}
\end{figure}

\cref{M_performance} shows the performance of GA and RS depending on the number of iterations ($M$). The performance was measured by taking the average of 100 test data solved by each method at each $M$. GA outperformed RS at every $M$, and the performance increased with increasing $M$ for both methods. However, the gradient of performance increment decreased with increasing $M$. On the other hand, our \ourmethod{} showed higher performance than GA$\{M=100\}$ and RS $\{M=10,000\}$ with a single inference $M=1$.
\vspace{5mm}
\section{Experimental Results in terms of Power Integrity}
\label{append: PI}

The objective of DPP is to suppress the probing port impedance as much as possible over a specified frequency range and is measured by the objective metric, $ Obj:=\sum_{f \in F} (Z_{initial}(f)-Z_{final}(f)) \cdot \frac{\text{1GHz}}{f}$. Performance of \ourmethod{} was evaluated in comparison to GA $\{M=100\}$ (\textit{expert policy}), GA $\{M=500\}$, RS $\{M=10,000\}$, AM-CRL and AM-CIL on unseen 100 PDN cases. Each method was asked to place 20 decaps ($K=20$) on each test. 

\newpage
\vspace{5mm}
\subsection{Impedance Suppression Plots}
\label{append: impedance-plots}
\vspace{-5mm}

\begin{figure}[h!]
    \centering
    \begin{subfigure}[b]{0.47\linewidth}
        \centering
        \includegraphics[width=\textwidth]{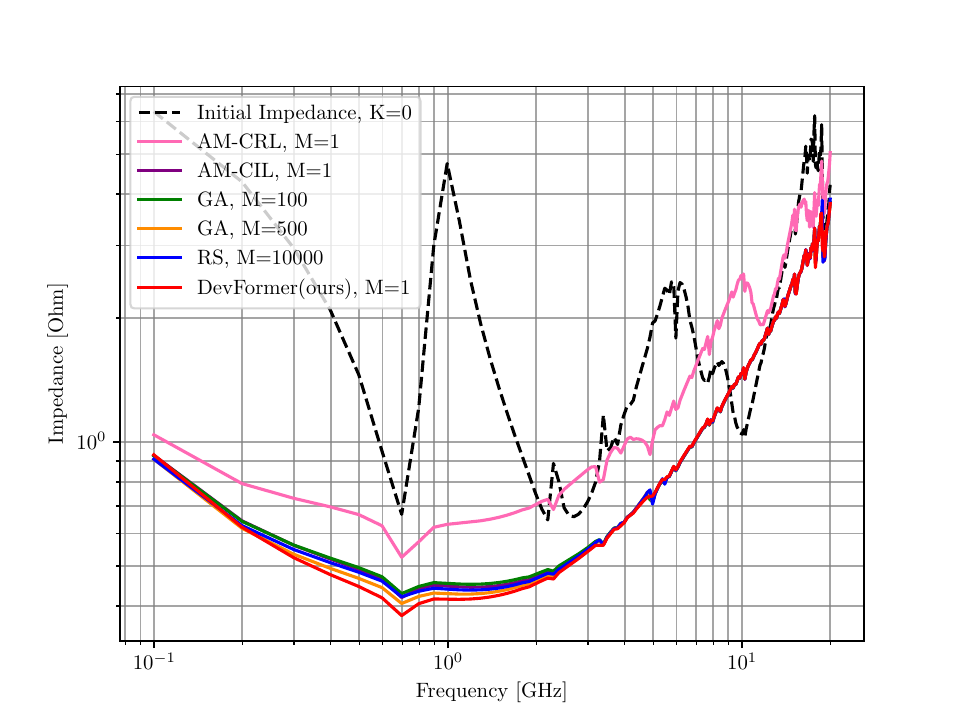}
        \caption{Test Case 1.}
    \end{subfigure}
    \begin{subfigure}[b]{0.47\linewidth}
        \centering
        \includegraphics[width=\textwidth]{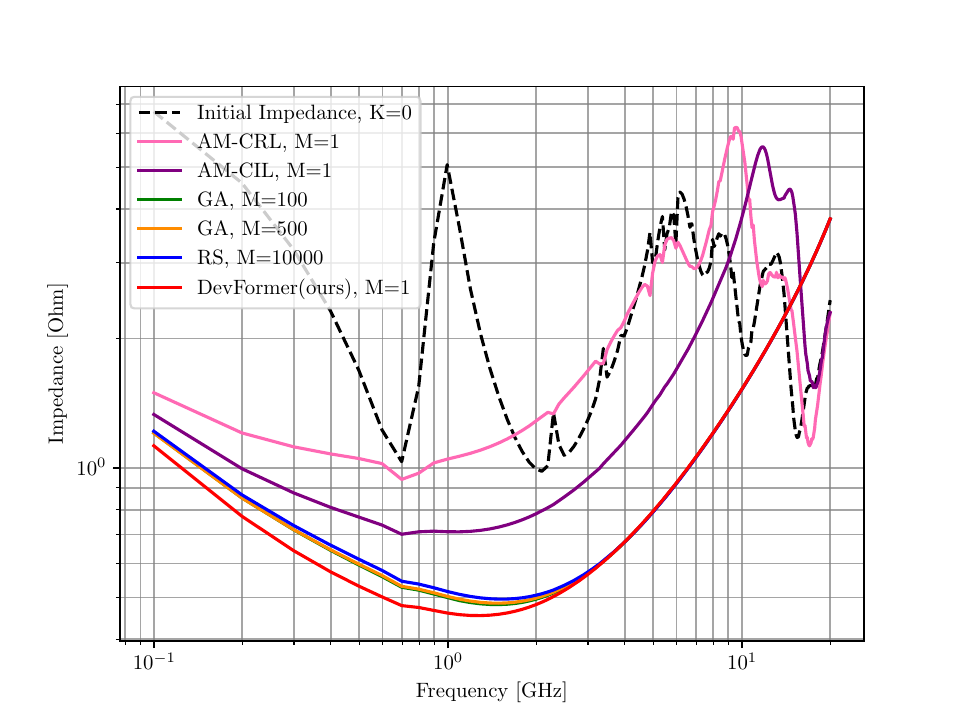}
        \caption{Test Case 2.}
    \end{subfigure}
    \begin{subfigure}[b]{0.47\linewidth}
        \centering
        \includegraphics[width=\textwidth]{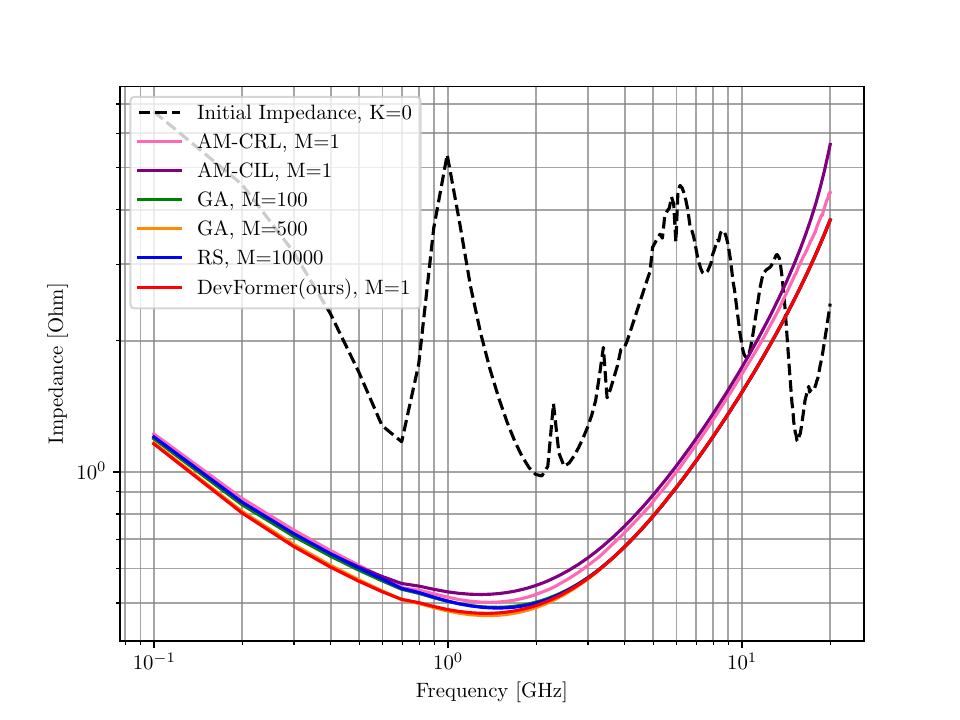}
        \caption{Test Case 3.}
    \end{subfigure}
    \begin{subfigure}[b]{0.47\linewidth}
        \centering
        \includegraphics[width=\textwidth]{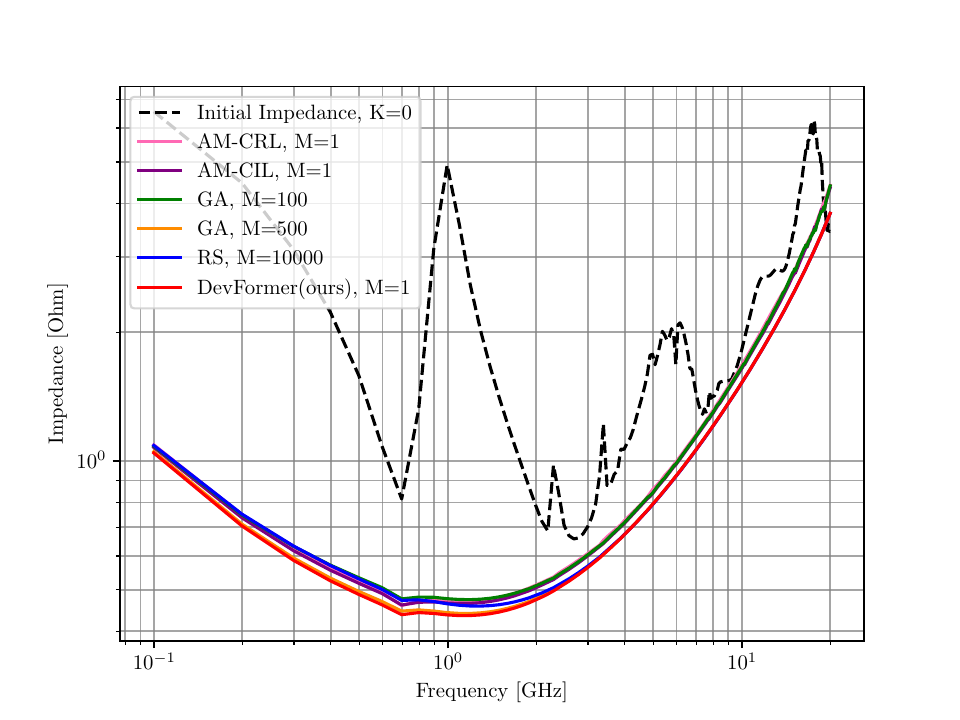}
        \caption{Test Case 4.}
    \end{subfigure}
        \begin{subfigure}[b]{0.47\linewidth}
        \centering
        \includegraphics[width=\textwidth]{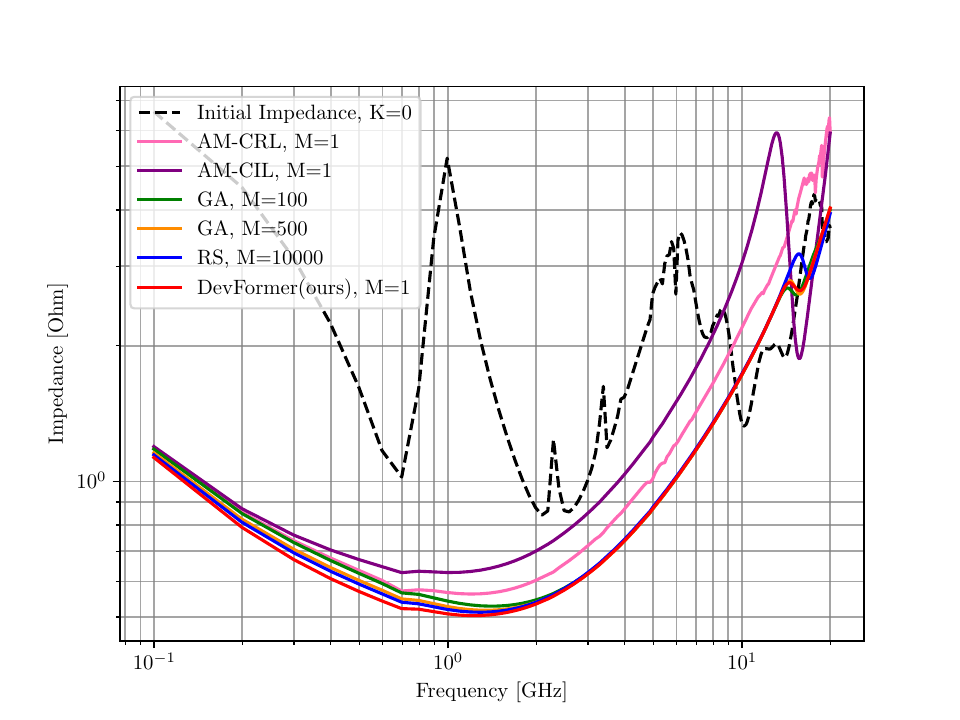}
        \caption{Test Case 5.}
    \end{subfigure}
        \begin{subfigure}[b]{0.47\linewidth}
        \centering
        \includegraphics[width=\textwidth]{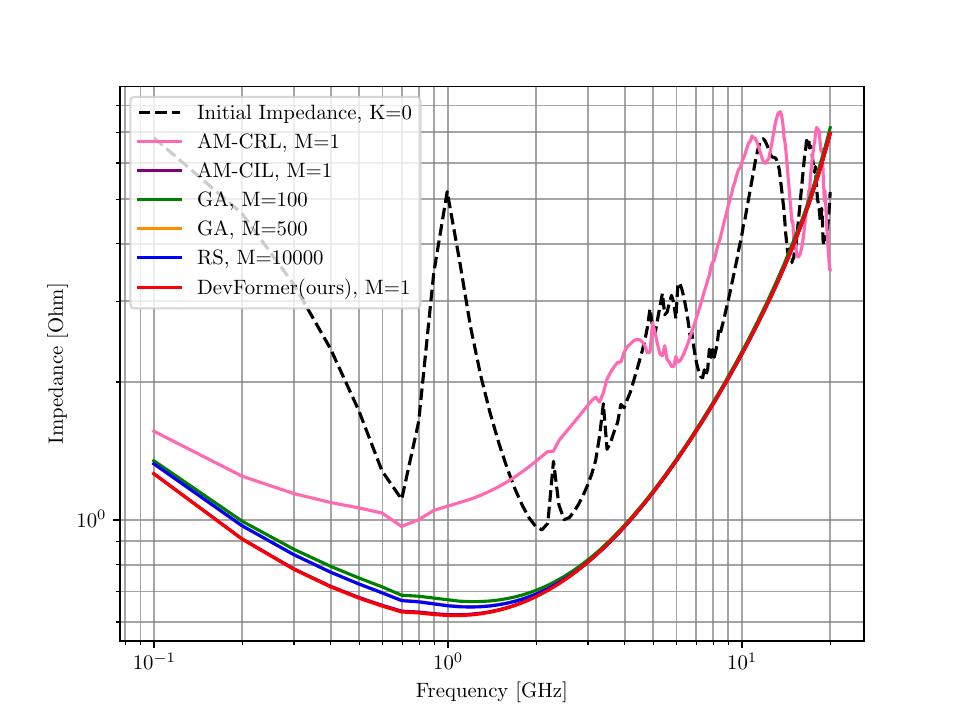}
        \caption{Test Case 6.}
    \end{subfigure}
    \caption{Impedance suppressed by each method, GA $\{M=100\}$ (\textit{expert policy}), GA $\{M=500\}$, RS $\{M=10,000\}$, AM-CRL, AM-CIL and \ourmethod{} (Ours) for 6 examples PDN cases out of 100 test dataset. (The lower, the better.)}
    \label{test_cases}
\end{figure}

\newpage
\subsection{Decap Placement Tendency Analysis}
\label{append: tendency}

\begin{figure}[h!]
    \centering
    \begin{subfigure}[b]{\linewidth}
        \centering
        \includegraphics[width=\textwidth]{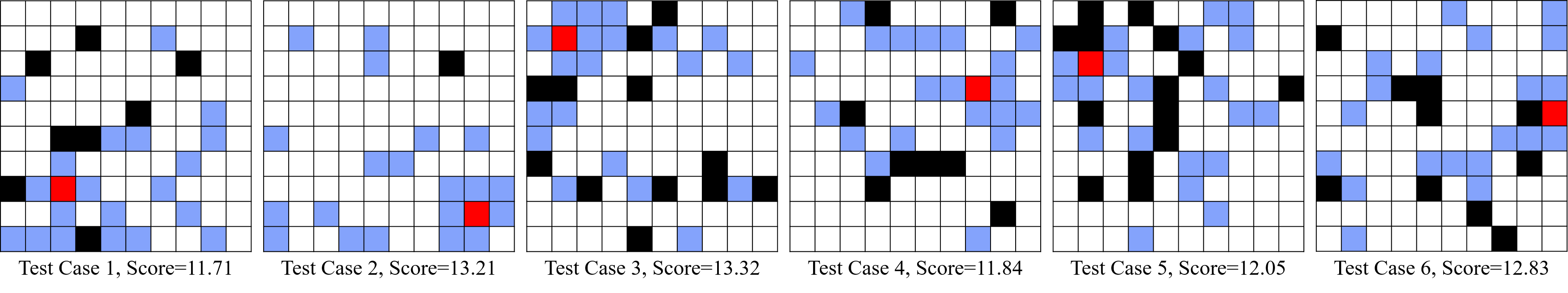}
        \vspace{-7mm}
        \caption{GA $\{M=100\}$.}
    \end{subfigure}
    \begin{subfigure}[b]{\linewidth}
        \centering
        \vspace{1mm}
        \includegraphics[width=\textwidth]{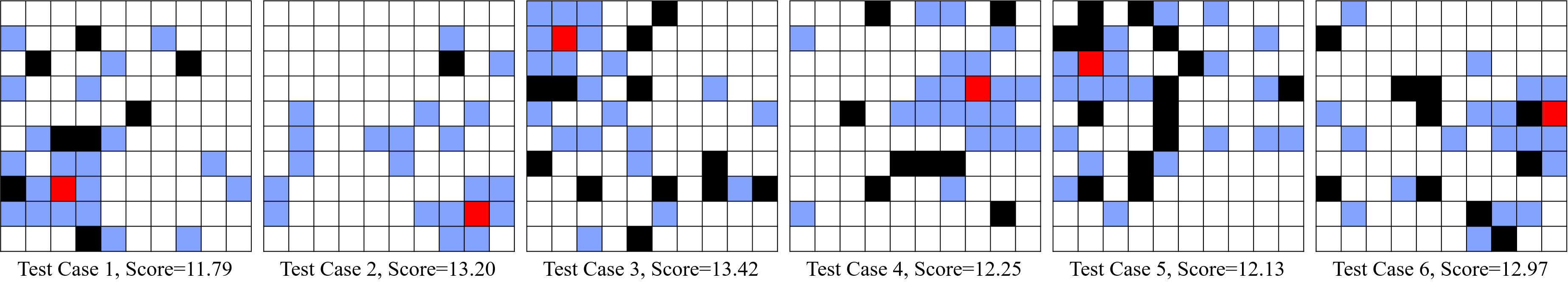}
        \vspace{-7mm}
        \caption{GA $\{M=500\}$.}
    \end{subfigure}
    \begin{subfigure}[b]{\linewidth}
        \centering
        \vspace{1mm}
        \includegraphics[width=\textwidth]{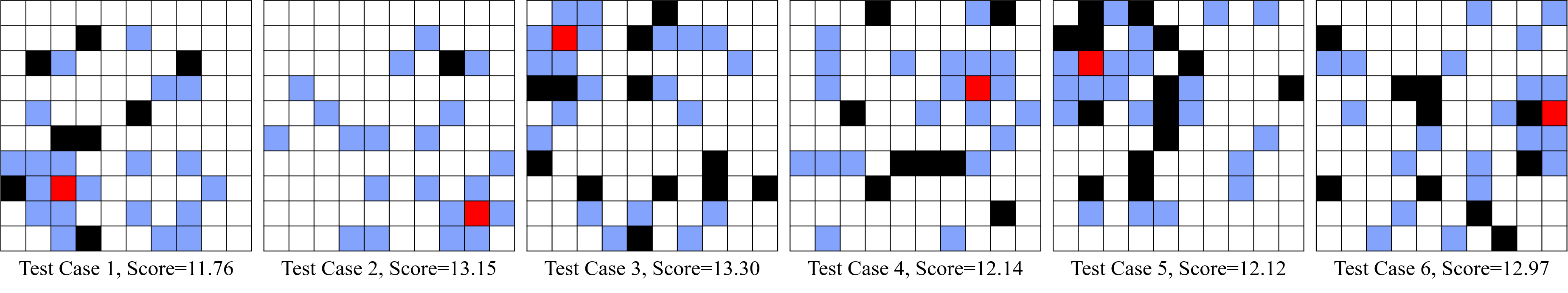}
        \vspace{-7mm}
        \caption{RS $\{M=10,000\}$.}
    \end{subfigure}
    \begin{subfigure}[b]{\linewidth}
        \centering
        \vspace{1mm}
        \includegraphics[width=\textwidth]{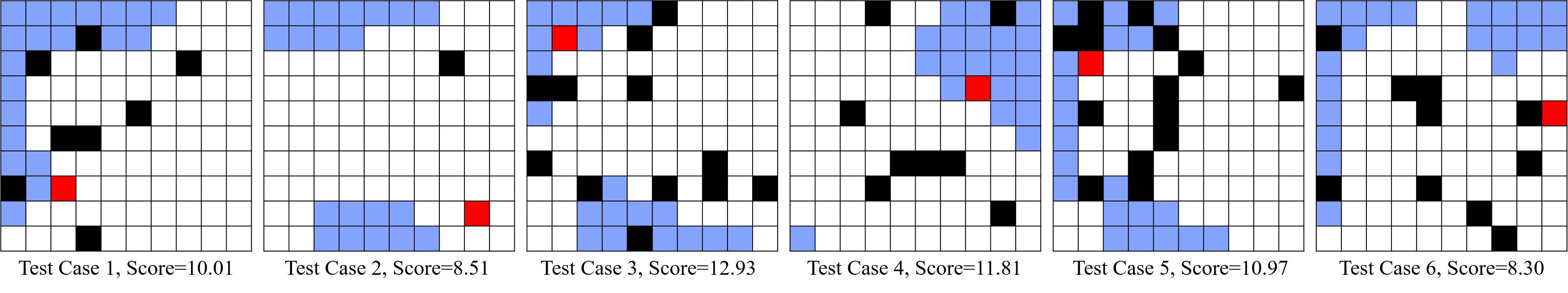}
        \vspace{-7mm}
        \caption{AM-CRL $\{M=1\}$.}
    \end{subfigure}
    \begin{subfigure}[b]{\linewidth}
        \centering
        \vspace{1mm}
        \includegraphics[width=\textwidth]{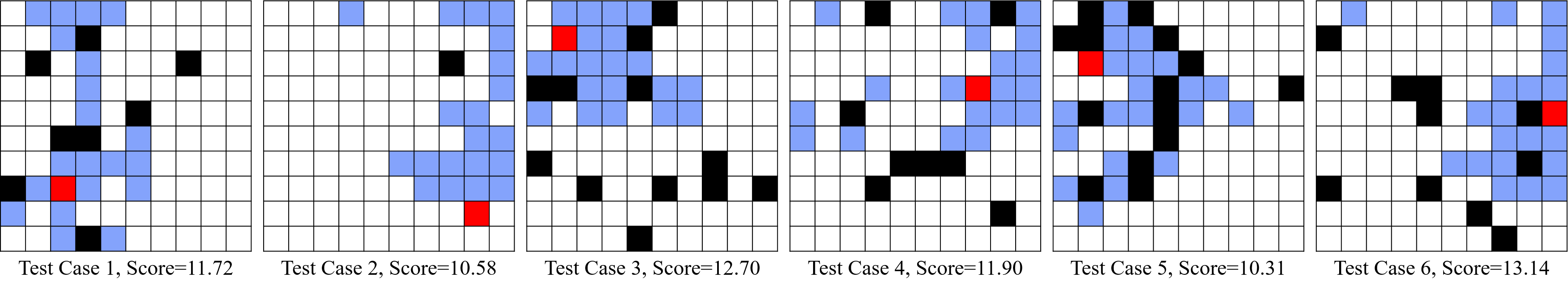}
        \vspace{-7mm}
        \caption{AM-CIL $\{M=1\}$.}
    \end{subfigure}
    \begin{subfigure}[b]{\linewidth}
        \centering
        \vspace{1mm}
        \includegraphics[width=\textwidth]{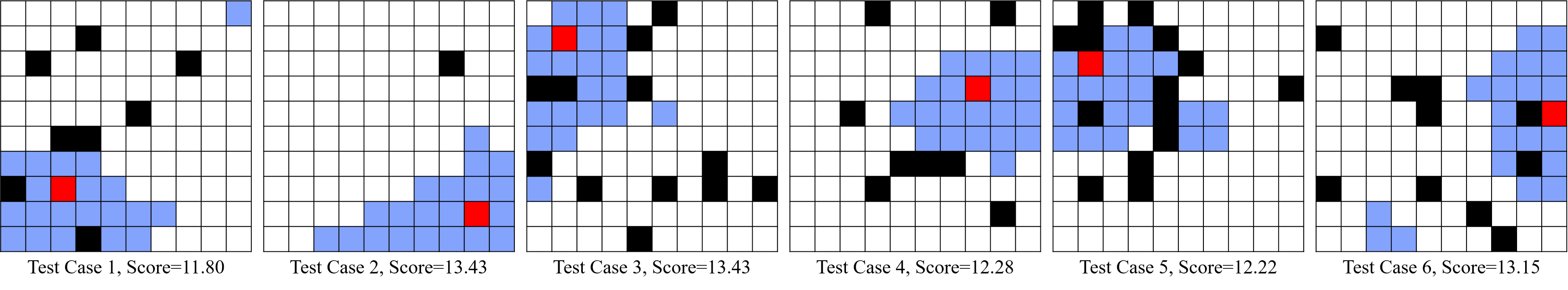}
        \vspace{-7mm}
        \caption{\ourmethod{}(ours)$\{M=1\}$.}
    \end{subfigure}
    \vspace{-7mm}
    \caption{Corresponding decap placement solutions to \cref{test_cases} by each method. Red represents probing port, black represents keep-out ports and blue represents decap locations.}
    \label{solution}
\end{figure}

\clearpage

\cref{solution} shows the decap placement solutions of 6 PDN cases plotted in \cref{test_cases}. The solutions by the search-heuristic methods, GA and RS, tend to be scattered, while the solutions by learning-based methods, AM-CRL, AM-CIL, and \ourmethod{}, are clustered. Since search-heuristic methods are based on random generations, they do not show a clear tendency. On the other hand, learning-based methods are based on a policy, having a distinct tendency to place decaps. 

The role of placing decaps in hardware design is to decouple the loop inductance of PDN. In terms of PI, analysis of loop inductance is critical, but at the same time, is complex \citep{Farrahi2019DesignCon2E}. The loop inductance distribution of PDN highly depends on various design parameters such as the location of the probing port, spacing between power/ground, size of PDN, and hierarchical layout of PDN \citep{loop_inductance}. When human experts place decaps on PDN, there are too many domain rules to consider. On the other hand, \ourmethod{} understands the PDN structure and its electrical properties by data-driven learning. According to \cref{solution}, \ourmethod{} tends to place decaps near the probing port, which is a well-known expert rule in the PI domain.

\subsection{Power Noise Analysis on HBM PDN}
\label{appendix:power_noise}
\vspace{-3mm}
\begin{figure}[h!]
    \centering
    \begin{subfigure}[b]{\linewidth}
        \includegraphics[width=0.97\textwidth]{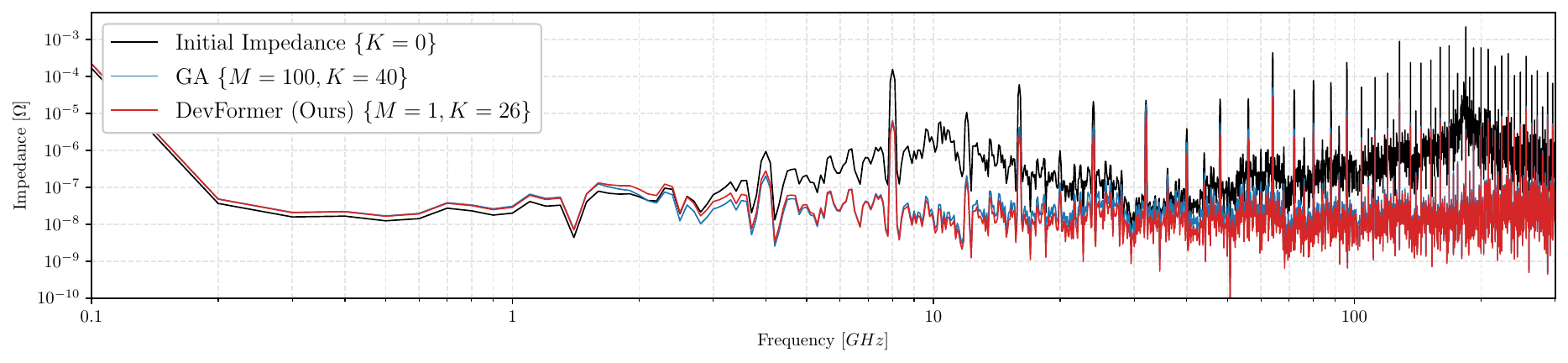}
        \vspace{-3mm}
        \caption{Impedance suppression}
        \label{SSN_A}
    \end{subfigure}
    \begin{subfigure}[b]{\linewidth}
        \centering
        \includegraphics[width=0.97\textwidth]{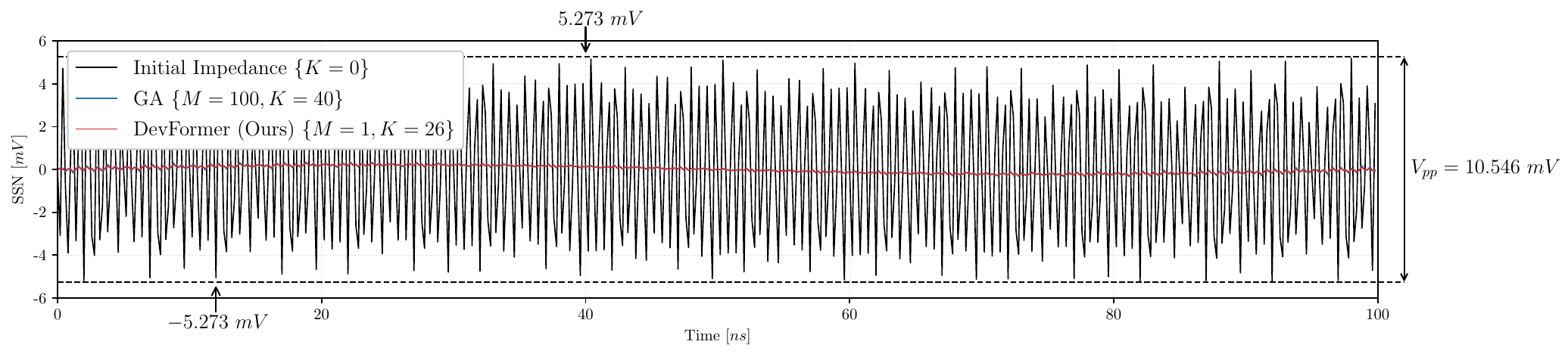}
        \vspace{-3mm}
        \caption{Initial power noise before decap placement}
        \label{SSN_B}
    \end{subfigure}
    \begin{subfigure}[b]{\linewidth}
        \centering
        \includegraphics[width=0.97\textwidth]{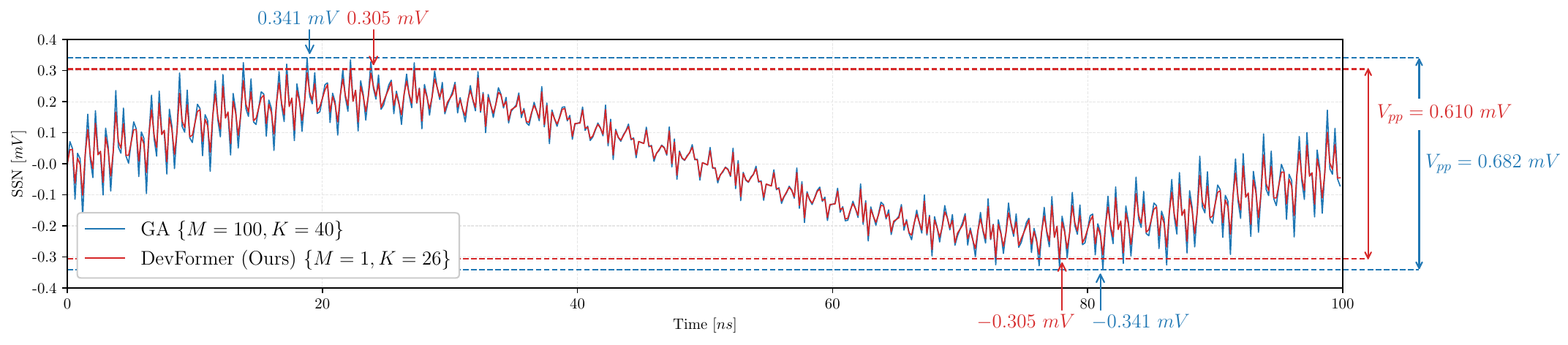}
        \vspace{-2mm}
        \caption{Power noise after decap placement by our \ourmethod{}$\{M=1, K=26\}$ and GA$\{M=100, K=40\}$} 
        \label{SSN_C}
    \end{subfigure}
    \vspace{-3mm}
    \caption{Power noise analysis in terms of simultaneous switching noise (SSN) on HBM PDN before and after decap placement by our \ourmethod{}$\{M=1, K=26\}$ and GA$\{M=100, K=40\}$. \ourmethod{} reduces power noise more than GA while reducing the needed decap number $K$ by more than $30\%$.}
    \label{solution}
\end{figure}
\vspace{-2mm}

\cref{appendix:power_noise} analyzes the performance of \ourmethod{} in comparison to GA\{$M=100$\} in terms of power noise. Out of 100 test cases on HBM PDN, we randomly chose a test case and carried out peak-to-peak power noise analysis for a circuit block, phase-locked loop (PLL), operating at 5GHz. Note that \ourmethod{} placed 26 decaps and GA\{$M=100$\} placed 40 decaps. \ourmethod{} reduced power noise more than GA\{$M=100$\} with 14 less decaps. The impedances of the probing port on the power distribution network (PDN) before and after decap placed by \ourmethod{} and GA\{$M=100$\} are presented in \cref{SSN_A}. The time-domain power noise before and after decap placement by each method is shown in \cref{SSN_B}. For performance comparison, \cref{SSN_C} shows the time-domain power noise after decap placement by \ourmethod{} and GA\{$M=100$\}.

\clearpage
\section{Further Ablation Study}
\label{append: ablation}

This section reports further ablation studies on the hyperparameters $N$ (number of offline expert data used), $\lambda$ (weight of self-exploitation loss term), and $P$ (number of permutation transformed labels).

\subsection{Ablation Study on $N$}
\label{append: N}

$N$ is the number of expert labels generated by the expert policy, GA $\{M=100\}$. We ablate $N \in \{100,500,1000,2000\}$ with fixed $P=4$ and $\lambda=5$ and compare to AM-CIL baseline for all $N$. As shown in \cref{table_N}, \ourmethod{} with $N=2000$ gives the best performance, and \ourmethod{} outperforms AM-CIL for all $N$ variations. The performance of AM-CIL is saturated at $N>500$ while the performance of \ourmethod{} continuously increases with the increase of $N$.

\begin{table}[h]
\fontsize{9}{9}\selectfont
\begin{center}

\caption{{Ablation study on $N$ for \ourmethod{} ($P=4, \lambda=8$) and AM-CIL.}}\label{table_N}
\vspace{1mm}
\begin{tabular}{lc}
\specialrule{1.0pt}{0pt}{4pt}
 & Validation Score \\
\specialrule{1.0pt}{0pt}{4pt}
AM-CIL \{$N=100$\}  & 11.02 \\ 
\ourmethod{} (ours) \{$N=100$\}& \textbf{12.76} \\
\specialrule{1.0pt}{0pt}{4pt}
AM-CIL \{$N=500$\}  & 11.80 \\ 
\ourmethod{} (ours) \{$N=500$\}& \textbf{12.85} \\
\specialrule{1.0pt}{0pt}{4pt}
AM-CIL \{$N=1000$\}  & 11.99 \\ 
\ourmethod{} (ours) \{$N=1000$\}& \textbf{12.86} \\
\specialrule{1.0pt}{0pt}{4pt}
AM-CIL \{$N=2000$\}  & 11.77 \\ 
\ourmethod{} (ours) \{$N=2000$\}& \textbf{12.88} \\
\specialrule{1.0pt}{0pt}{4pt}
\label{table_N}
\end{tabular}
\end{center}
\end{table}

\begin{figure}[h]
\centerline{\includegraphics[width=0.6\textwidth]{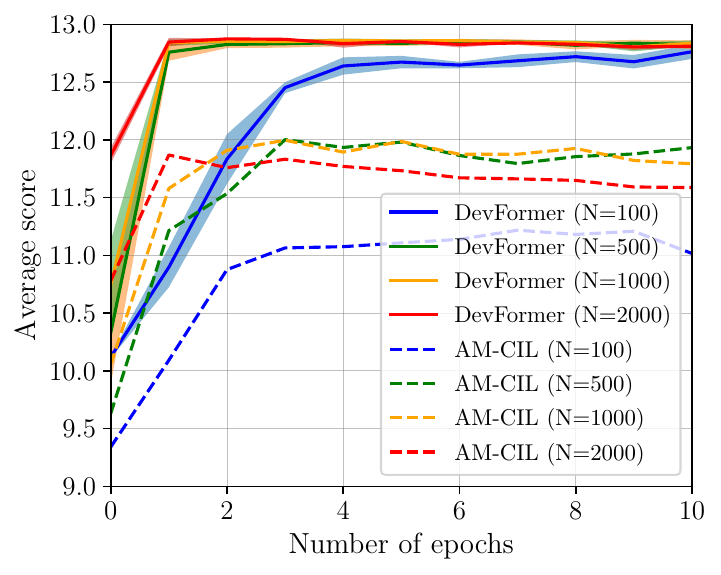}}
\caption{Validation graph of \ourmethod{} and AM-CIL for varying number of offline expert data $N \in \{100, 500, 1000, 2000\}$.}
\label{figure_N}
\end{figure}

\clearpage
\subsection{Ablation Study on $\lambda$}
\label{append: lambda}

$\lambda$ refers to the weight of self-exploitation loss term $L_{Self}$, in the collaborative learning loss $\mathcal{L} := \mathcal{L}_{Expert} + \lambda \mathcal{L}_{Self}$. To set $\lambda \times L_U$ to be $0.1 \sim 1$, we first multiplied $10^{32}$ to $\lambda$ because the probability of a specific solution is extremely small. Then, we ablated for $\lambda \in \{1,2,4,6,7,8,9,10 \}$ ($10^{32}$ is omitted) with fixed $N=2000$ and $P=4$. For every $\lambda$, it prevents overfitting of the model in comparison to the baselines trained only with $L_{Expert}$ (see \cref{figure_lambda}). According to the \cref{table_lambda}, $\lambda = 8$ gives the best validation scores.  

\begin{table}[h]
\fontsize{9}{9}\selectfont
\begin{center}
\caption{{Ablation study of $\lambda$ on fixed $P=4$ and $N=2000$.}}\label{table_lambda}
\vspace{1mm}
\begin{tabular}{lc}
\specialrule{1.0pt}{0pt}{4pt}
$\lambda$ ($\times 10^{32}$) & Validation Score \\
\specialrule{0.5pt}{0pt}{4pt}
1 & 12.863 \\
2 & 12.865 \\
3 & 12.866\\
4 & 12.870\\
\textbf{5} & \textbf{12.877}\\
6 & 12.874\\
7 & 12.862\\
8 & 12.871 \\
9 & 12.871 \\
10 & 12.870 \\
\specialrule{0.5pt}{0pt}{4pt}
Only IL & 11.832 \\
\specialrule{1.0pt}{0pt}{4pt}
\label{table_lambda}
\end{tabular}
\end{center}
\end{table}

\begin{figure}[h]
\centerline{\includegraphics[width=0.5\textwidth]{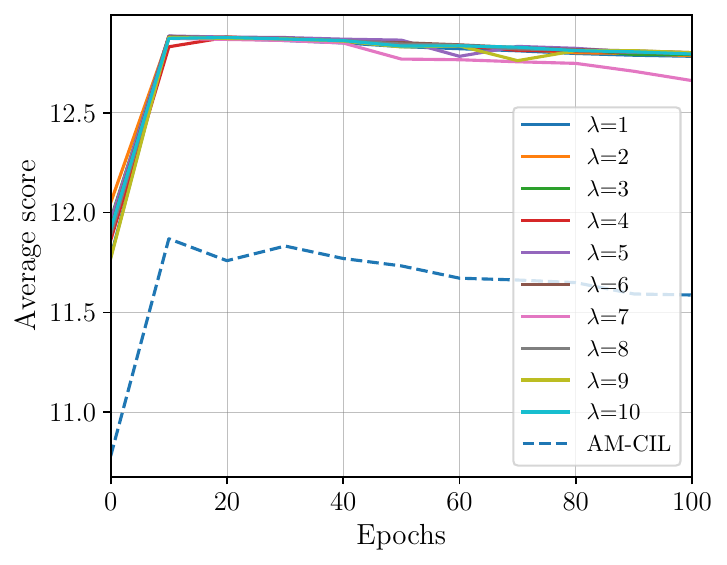}}
\caption{Validation graph of $\lambda \in \{1,2,3, 4,5, 6,7,8,9,10\}$ on fixed $P=4$ and $N=2000$.}
\label{figure_lambda}
\end{figure}

\clearpage

\subsection{Ablation Study on $P$}
\label{append: P}

$P$ is the number of permutation-transformed labels per each expert label used for imitation learning-based expert exploitation. We ablate $P \in \{4,6,8\}$ with fixed $N=2000$ and $\lambda=5$ and compared collaborative symmetricity exploitation (i.e., both expert and self-exploitation) to only expert exploitation training cases. As shown in \cref{table_P}, $P=4$ with \{Expert exploitation + Self-exploitation\} give best performances. For every $P$, \{Expert exploitation + Self-exploitation\} gives better performances, indicating the self-exploitation scheme well prevents overfitting the training process for the sparse dataset.

\begin{table}[h]
\fontsize{9}{9}\selectfont
\begin{center}
\caption{{Ablation study on $P$ with and without unsupervised loss term.}}\label{table_P}
\vspace{1mm}
\begin{tabular}{lc}
\specialrule{1.0pt}{0pt}{4pt}
 & Validation Score \\
\specialrule{1.0pt}{0pt}{4pt}
Expert exploitation \{$P=4$\}  & 12.85 \\ 
+ Self- exploitation \{$\lambda = 5$\}& \textbf{12.88} \\
\specialrule{1.0pt}{0pt}{4pt}
Expert exploitation \{$P=6$\} & 12.87 \\
+ Self- exploitation \{$\lambda = 5$\}& \textbf{12.88} \\
\specialrule{1.0pt}{0pt}{4pt}
Expert exploitation \{$P=8$\} & 12.88 \\
+ Self- exploitation \{$\lambda = 5$\}& \textbf{12.88} \\
\specialrule{1.0pt}{0pt}{4pt}
\label{table_lambda}
\end{tabular}
\end{center}
\end{table}

\begin{figure}[h]
\centerline{\includegraphics[width=0.6\textwidth]{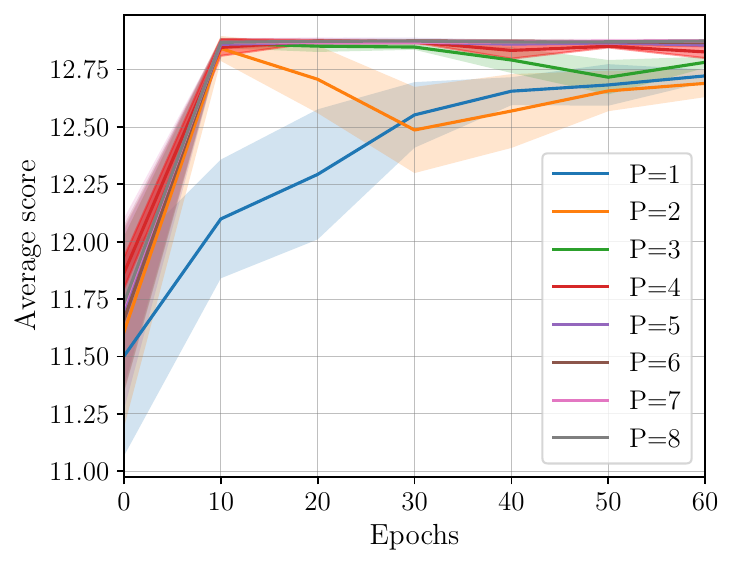}}
\caption{Validation score of $P$ ablation with and without self-exploitation loss term.}
\label{figure_P}
\end{figure}

\clearpage
\section{Proof of \cref{thm:order-bias}}
\label{append:proof}

$(\rightarrow)$ Suppose that policy $\pi(\boldsymbol{a}|\boldsymbol{x})$ is AP-symmetric. Then, by the \cref{def:ap-sym}, $\pi(\boldsymbol{a}|\boldsymbol{x}) = \pi(t(\boldsymbol{a})|\boldsymbol{x})$ for any $\boldsymbol{a} \in \mathcal{A}$, $\boldsymbol{x} \in \mathcal{X}$, $t \in T_{AP}$. 

Therefore,
\begin{equation*}
    b(\pi;\boldsymbol{p}) = \mathbb{E}_{\boldsymbol{x} \sim  p_{\mathcal{X}}(x)}\mathbb{E}_{\boldsymbol{a} \sim p_{\mathcal{A}}(\boldsymbol{a})}\mathbb{E}_{t \sim p_{T_{AP}}(t)}[||\pi(\boldsymbol{a}|\boldsymbol{x})-\pi(t(\boldsymbol{a})|\boldsymbol{x})||_{1}] = 0 
\end{equation*}

$(\leftarrow)$ Suppose that $b(\pi;\boldsymbol{p})=0$, where $p_{\mathcal{X}}(\boldsymbol{x})>0, p_{\mathcal{A}}(\boldsymbol{a})>0, p_{T_{AP}}(t)>0$.

Assume that there exist $\boldsymbol{a^{*}} \in \mathcal{A}$, $\boldsymbol{x^{*}} \in \mathcal{X}$, and $t^{*} \in T_{AP}$, such that $\pi(\boldsymbol{a}^{*}|\boldsymbol{x}^{*}) \neq \pi(t(\boldsymbol{a}^{*})|\boldsymbol{x}^{*})$.

Then, 
\begin{align*}
b(\pi;\boldsymbol{p}) = \mathbb{E}_{\boldsymbol{x} \sim  p_{\mathcal{X}}(x)}\mathbb{E}_{\boldsymbol{a} \sim p_{\mathcal{A}}(\boldsymbol{a})}\mathbb{E}_{t \sim p_{T_{AP}}(t)}[||\pi(\boldsymbol{a}|\boldsymbol{x})-\pi(t(\boldsymbol{a})|\boldsymbol{x})||_{1}] \\\geq p_{\mathcal{X}}(\boldsymbol{x}^{*})p_{\mathcal{A}}(\boldsymbol{a}^{*})p_{T_{AP}}(t^{*})||\pi(\boldsymbol{a}^{*}|\boldsymbol{x}^{*})-\pi(t(\boldsymbol{a}^{*})|\boldsymbol{x}^{*})||_{1}>0,
\end{align*}

which results in a contradiction. Therefore, $\pi(\boldsymbol{a}|\boldsymbol{x}) = \pi(t(\boldsymbol{a})|\boldsymbol{x})$ for any F$\boldsymbol{a} \in \mathcal{A}$, $\boldsymbol{x} \in \mathcal{X}$, $t \in T_{AP}$: i.e, policy $\pi(\boldsymbol{a}|\boldsymbol{x})$ is AP-symmetric.


\end{document}